\documentclass[letterpaper]{article} %
\usepackage{aaai2026}  %
\usepackage{times}  %
\usepackage{helvet}  %
\usepackage{booktabs} %
\usepackage{multirow}
\usepackage{placeins}
\usepackage{courier}  %
\usepackage[hyphens]{url}  %
\usepackage{graphicx} %
\usepackage{natbib}  %
\usepackage{caption} %
\usepackage{etoolbox}

\AtBeginEnvironment{algorithm}{\setlength{\parindent}{0pt}}

\usepackage{algorithm}
\usepackage{algorithmicx}
\usepackage{algpseudocode}
\usepackage{amssymb}
\usepackage{amsmath}
\usepackage{tcolorbox}
\usepackage{algorithm}
\usepackage{algpseudocode}

\usepackage[T1]{fontenc}
\usepackage{booktabs,multirow}
\usepackage{makecell}
\usepackage{tabulary}
\usepackage{fontawesome5}
\newcommand \blfootnote[1]{
    \begingroup
        \renewcommand
        \thefootnote{}\footnote{#1}
        \addtocounter{footnote}{-1}
        \vspace{-1ex}
    \endgroup
}
\newlength\savewidth\newcommand\shline{\noalign{\global\savewidth\arrayrulewidth
  \global\arrayrulewidth 1pt}\hline\noalign{\global\arrayrulewidth\savewidth}}

\newcommand{\tablestyle}[2]{\setlength{\tabcolsep}{#1}\renewcommand{\arraystretch}{#2}\centering\footnotesize}

\usepackage{pifont}
\usepackage{tikz}

\usepackage{bbding}
\usepackage{cleveref}

\usepackage{newfloat}
\usepackage{listings}
\DeclareCaptionStyle{ruled}{labelfont=normalfont,labelsep=colon,strut=off} %
\floatstyle{ruled}
\newfloat{listing}{tb}{lst}{}
\floatname{listing}{Listing}
\newcommand{\para}[1]{\noindent\textbf{#1}}
\usepackage[dvipsnames]{xcolor}
\newcommand{\mytiny}[1]{\scriptsize{#1}}
\renewcommand\shline{\noalign{\global\savewidth\arrayrulewidth
  \global\arrayrulewidth 1pt}\hline\noalign{\global\arrayrulewidth\savewidth}}

\title{Filter, Correlate, Compress: \\ Training-Free Token Reduction for MLLM Acceleration}

\author{
    Yuhang Han\textsuperscript{\rm 1*},
    Xuyang Liu\textsuperscript{\rm 2*},
    Zihan Zhang\textsuperscript{\rm 3},
    Pengxiang Ding\textsuperscript{\rm 1},
    Junjie Chen\textsuperscript{\rm 2}, \\
    Donglin Wang\textsuperscript{\rm 1},
    Honggang Chen\textsuperscript{\rm 2},
    Qingsen Yang\textsuperscript{\rm 4,5},
    Siteng Huang\textsuperscript{\rm 6\textsuperscript{\dag}} \\ %
}

\affiliations{
    \textsuperscript{\rm 1}Westlake University,
    \textsuperscript{\rm 2}Sichuan University,
    \textsuperscript{\rm 3}Johns Hopkins University, 
    \textsuperscript{\rm 4}Northwestern Polytechnical University, \\
    \textsuperscript{\rm 5}Shenzhen Research Institute of Northwestern Polytechnical University,
    \textsuperscript{\rm 6}Zhejiang University \\
    yuhangh984@gmail.com, siteng.huang@gmail.com
}

\usepackage{bibentry}

\begin{document}

\maketitle
\blfootnote{* Equal contribution.}
\blfootnote{\textsuperscript{\dag} Corresponding author.}

\begin{abstract}

The quadratic complexity of Multimodal Large Language Models (MLLMs) with respect to context length poses significant computational and memory challenges, hindering their real-world deployment.
In the paper, we devise a ``\textbf{\textit{filter-correlate-compress}}'' framework to accelerate the MLLM by systematically optimizing multimodal context length during prefilling. The framework first implements \textbf{\textit{FiCoCo-V}}, a training-free method operating within the vision encoder.
It employs a redundancy-based token discard mechanism that uses a novel integrated metric to accurately \textit{filter} out redundant visual tokens.
To mitigate information loss, the framework introduces a correlation-based information recycling mechanism that allows preserved tokens to selectively recycle information from \textit{correlate}d discarded tokens with a self-preserving \textit{compress}ion, thereby preventing the dilution of their own core content. The framework's \textbf{\textit{FiCoCo-L}} variant further leverages task-aware textual priors to perform token reduction directly within the LLM decoder. Extensive experiments demonstrate that the \textit{FiCoCo} series effectively accelerates a range of MLLMs, achieves up to \textbf{14.7×} FLOPs reduction with \textbf{93.6\%} performance retention. Our methods consistently outperform state-of-the-art training-free approaches, showcasing effectiveness and generalizability across model architectures, sizes, and tasks without requiring retraining.

\end{abstract}

\begin{links}
    \link{Code}{https://github.com/kawhiiiileo/FiCoCo}
\end{links}    
\section{Introduction}
\label{sec:intro}

\begin{figure}[!t]
  \centering
  \includegraphics[width=\linewidth]{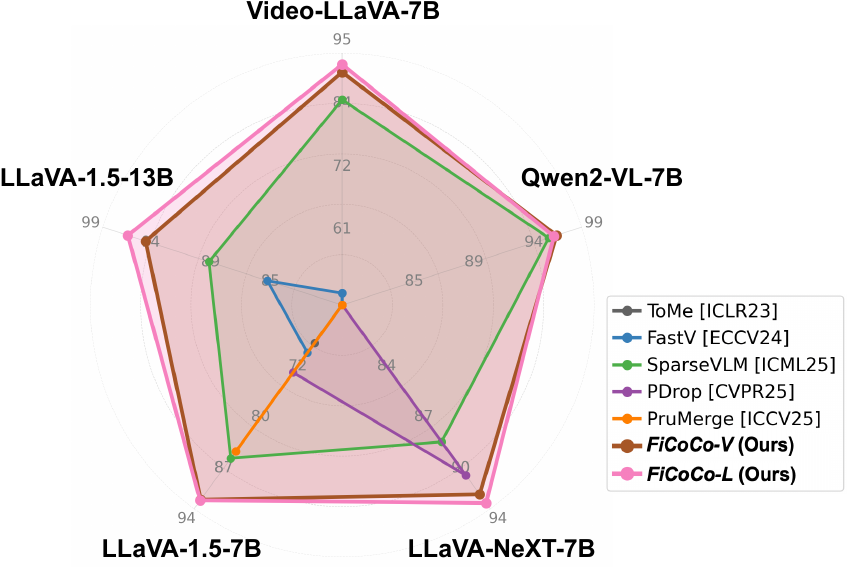}
  \caption{
  \textbf{The comparison to existing token reduction methods.}
  Our \textit{FiCoCo} series achieves state-of-the-art results with five popular MLLMs across benchmarks. %
  }
  \label{fig:acc_with_flops}
\end{figure}

Multimodal Large Language Models (MLLMs)~\cite{Liu:LLaVA,Liu:LLaVA-1.5,Zhang:Video-LLaMA,Chen:InternVL,Lin:VILA} have effectively extended the impressive emergent capabilities of Large Language Model (LLM)~\cite{Touvron:LLaMA,OpenAI:GPT-4,Bai:Qwen} decoders by integrating visual features with textual inputs.
However, the substantial increase in context length when processing images and videos imposes quadratically scaling computational and memory demands.
This, in turn, renders prefilling a critical bottleneck for MLLM response generation.
As an empirical instance, the prefilling time for Qwen2-VL~\cite{Wang:Qwen2-VL} in high-resolution visual question answering significantly outweighs its decoding time, constituting up to a remarkable \textbf{80\%} of the total latency.
In this paper, we introduce ``\textbf{\textit{filter-correlate-compress}}'', a framework that systematically and progressively optimizes the length of multimodal context during prefilling.
This enables MLLMs to minimize response latency while concurrently striving for the preservation of generation quality.

Concurrently, we propose \textbf{\textit{FiCoCo-V}}, a training-free method that represents the framework's implementation within the vision encoder, primarily addressing data redundancy.
Natural vision signals, such as images and videos, inherently possess a higher degree of information redundancy compared to human-generated languages~\cite{He:MAE,Feichtenhofer:MEA-Video}.
However, in the constructed multimodal context, the number of visual tokens substantially exceeds that of textual tokens.
Our framework thus initiates with a \textbf{redundancy-based token discard}, which reduces context length by measuring and \textbf{\textit{filter}}ing out redundant visual tokens at each layer.
Specifically, unlike a potentially biased, single redundancy metric~\cite{Chen:FastV,Zhang:SparseVLM}, \textit{FiCoCo-V} integrates vision-aware and semantic-aware redundancy to accurately discard those more redundant tokens.

One common oversight is that tokens considered redundant may contain noise or still hold information beneficial to the task~\cite{Liang:EViT}.
Developing effective mechanisms to flexibly recover such information facilitates performance maintenance of the MLLM.
However, preserved tokens must carefully select which information to receive from discarded ones, preventing excessive dilution of their core information.
We highlight that inter-token correlation provides a principled metric to guide where and how information should be recycled.
Moving forward, our framework designs a \textbf{correlation-based information recycling} mechanism that allows each redundant token to \textbf{\textit{correlate}} a variable number of preserved tokens to adaptively retain its information,
while the \textit{FiCoCo-V} method models such correlation with direct attention.
Subsequently, a self-preserving \textbf{\textit{compress}}ion operation ensures the prominence of the preserved tokens while allowing them to receive more information from highly correlated redundant tokens.

While \textit{FiCoCo-V} strikes an appealing balance between efficiency and performance, executing token discard and information recycling in a task-agnostic manner can be shortsighted, as it fundamentally constrains the performance ceiling of the acceleration method.
Consequently, we further propose \textbf{\textit{FiCoCo-L}}, which performs token reduction directly within the LLM decoder.
By leveraging task-aware textual priors, 
\textit{FiCoCo-L} more precisely pinpoints redundant tokens and recovers crucial information, thereby minimizing loss during the compression process.
In Figure~\ref{fig:acc_with_flops}, when applied to LLaVA-1.5-7B~\cite{Liu:LLaVA-1.5}, both methods consistently outperform existing token reduction baselines across different FLOPs.
In the most extreme case, our method can obtain a maximum improvement of \textbf{5.7$\times$} in FLOPs while retaining \textbf{92.8\%} performance.
When applied to the more powerful LLaVA-NeXT-7B~\cite{Liu:LLaVA-NeXT}, our methods even show stronger superiority, achieving a \textbf{14.7$\times$} improvement in FLOPs while retaining at most \textbf{93.6\%} performance.
We also evaluate our methods on video understanding tasks, where our methods retain at most \textbf{92.8\%} performance of vanilla Video-LLaVA~\cite{Lin:Video-LLaVA} with a \textbf{11.4$\times$} improvement in FLOPs.
As a conclusion, our success in token budget reduction and model acceleration can generalize across various MLLM architectures, sizes, and tasks.

\section{Related Work}

\para{Multimodal Large Language Models (MLLMs).}
To acquire visual comprehension and reasoning capabilities, MLLMs~\cite{Dai:InstructBLIP,Bai:Qwen-VL,Liu:LLaVA,Chen:InternVL} first use a pre-trained vision encoder (\textit{e.g.}, from CLIP~\cite{Radford:CLIP}) to extract visual features, which are then projected into the input embedding space of a pre-trained Large Language Model (LLM)~\cite{Touvron:LLaMA,OpenAI:GPT-4,Bai:Qwen} decoder.
The LLM then processes these visual embeddings alongside user instructions to understand the images and craft suitable responses.
A key trend in the development of MLLMs is to leverage longer multimodal contexts to capture finer-grained visual details, thereby enabling a more profound comprehension of the visual content.
For example, 
LLaVA-1.5~\cite{Liu:LLaVA-1.5} improves the vision encoder for higher resolutions, while
LLaVA-NeXT~\cite{Liu:LLaVA-NeXT} quadruples input resolution with flexible aspect ratios to enhance fine-grained understanding.
And Video-LLaVA~\cite{Zhang:Video-LLaMA} employs extended context windows and dynamic frame aggregation to accommodate longer input sequences for video-text tasks.
However, increased context length introduces significant inference latency and storage overheads, which become major deployment bottlenecks.

\para{Token Reduction for MLLMs.}    \label{sec:related-work-token-reduction}
Token reduction approaches can be broadly unified as token compression, which aims to eliminate redundancy and condense visual information into a more compact representation 
while minimizing information loss~\cite{Rao:DynamicViT,Liang:EViT,Bolya:ToMe,liu2025shifting,liu2025global,liu2025video,liu2025mixing}.
Our proposed methods adaptively adjust the number of tokens each discarded token is compressed into.
This functions as an automatic, per-token switching mechanism between the two techniques, designed to maximize benefits.

Token reduction for MLLMs has gradually shifted from training-based methods~\cite{Cha:Honeybee,Li:TokenPacker} to training-free approaches~\cite{Chen:FastV,Zhang:SparseVLM}, as the latter enables direct application to off-the-shelf models without costly retraining overheads.
For instance,
FastV~\cite{Chen:FastV} prunes unnecessary visual tokens based on the ranking of attention scores derived from the self-attention mechanism in the LLM.
SparseVLM~\cite{Zhang:SparseVLM} adaptively prunes visual tokens in the LLM based on their attention scores with text tokens.
PDrop~\cite{Xing:PyramidDrop} drops visual tokens according to the attention between all the visual tokens and the last token of the instruction.
In this study, 
our \textit{FiCoCo} shows that more precise identification of redundant tokens and controlled recovery of discarded information can achieve superior performance while maintaining high efficiency.

\section{Methodology of \textit{FiCoCo-V}}\label{method}

\begin{figure*}[!t]
  \centering
   \includegraphics[width=\linewidth]{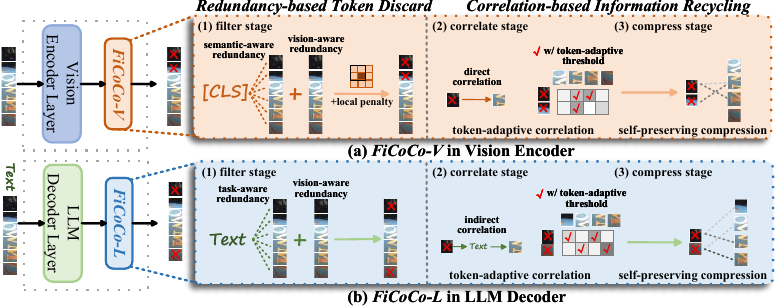}
   \caption{
   \textbf{Overview of \textit{FiCoCo-V} and \textit{FiCoCo-L}.}
   Due to the two methods being applied to different modules (vision encoder and LLM decoder), they have different implementations for summarized redundancy and correlation matrix in the filter and correlate stages. Simultaneously, the compression modules of \textit{FiCoCo-V} and \textit{FiCoCo-L} are identical, both employing self-preserving compression based on the correlation matrix.
   }
   \label{fig:method_example}
\end{figure*}

\subsection{Preliminaries: Revisiting MLLMs}    \label{sec:preliminaries}

\para{Prefilling.}
Given the textual instruction, a MLLM generates responses according to the input image, where the critical prefilling phase involves two key steps:
(1) Input tokenization, where the visual encoder extracts the visual features and projects them into a shared latent space with discrete textual tokens.
(2) Causal self-attention computation, wherein the LLM decoder performs causal self-attention over this entire concatenated sequence to establish contextual dependencies, providing intermediate key-value pairs for efficient autoregressive decoding.
This phase establishes the decoding context for subsequent token generation, directly impacting inference latency and throughput.  %

\para{Self-Attention.}
The self-attention mechanism~\cite{Vaswani:Transformer} stands as both the most essential and the most resource-intensive operation in transformer-based visual encoder and LLM decoder.
Given the input 1D sequence $\mathbf{X}$ of length $N$, the self-attention layer produces a self-attention map $\mathbf{A}\in \mathbb{R}^{N \times N}$ to globally model the dependence relationships between tokens, formulated as
$\mathbf{A}=\text{Attention}\left( \mathbf{Q}, \mathbf{K}\right)=\text{Softmax}\left( \mathbf{Q}{\mathbf{K}}^\top / \sqrt{D} \right),$
where $^\top$ denotes the transpose of the matrix, the query and key matrices $\mathbf{Q},\mathbf{K} \in \mathbb{R}^{N \times D}$ are obtained by projecting $\mathbf{X}$ with learnable parameter matrices.   %

\subsection{Redundancy-based Token Discard}

\para{Filter: What token should be discarded?}
When evaluating the redundancy of tokens within the vision encoder, we draw inspiration from the two natural principles of humans when quickly and comprehensively summarizing the content of a given image.
Firstly, to accelerate the recognition, we tend to ignore those similar pixels as they commonly provide the same information.
Similarly, within a self-attention layer, if a visual token requires substantial information from other visual tokens, it indicates that its own information is not unique, and the token can be replaced by other visual tokens.
Formally, given the self-attention weight matrix $\mathbf{A}^v \in \mathbb{R}^{N \times N}$, where $N$ is the number of the visual tokens, we can define the \textbf{vision-aware redundancy} of the $i$-th token by averaging its received attention, \textit{i.e.}, $\frac{1}{N} \sum_{j=1}^N \mathbf{A}^v_{i,j}$.
We emphasize that this design is significantly different from previous methods~\cite{Chen:FastV}, as they regard attention between visual tokens as a measure of importance, and provide analysis in Appendix.  %

Secondly, if provided with the overall concept of the image, we rapidly identify the area of interest based on the global semantic clue and ignore other regions.
As typical vision encoders~\cite{Dosovitskiy:ViT,Radford:CLIP} employ a \texttt{[CLS]} token to capture the global image representation, its attention weights $\mathbf{a}^{\texttt{CLS}}$ can quantify the global semantic content of patch tokens, which can be useful for multimodal understanding.
And we can define the \textbf{semantic-aware redundancy} by applying a \textit{negation} operation.
We regard this general solution as the default due to its efficiency, and provide an alternative solution for a limited number of MLLMs 
without a \texttt{[CLS]} token (\textit{e.g.}, SigLIP~\cite{Zhai:SigLIP}).
Specifically, we average the keys of all visual tokens as an alternative of the \texttt{[CLS]} token, and regard its cosine similarity with visual tokens as a substitute for attention.
Details and experiments of this alternative can be found in Appendix Table~\ref{tab:non-cls}.  %
Therefore, we calculate the overall redundancy score for each visual token as
{\setlength\abovedisplayskip{2mm}
\setlength\belowdisplayskip{0mm}
\begin{equation}\label{eq3}
\mathbf{s}^v_i = \lambda \frac{1}{N} \sum_{j=1}^N \mathbf{A}^v_{i,j} - (1 - \lambda)  \mathbf{a}_i^{\texttt{CLS}},    %
\end{equation}
}%
where $\lambda$ is a scalar hyperparameter that balances the factors.
Since the visual tokens with \textit{higher} redundancy scores are expected to be discarded, we filter out these tokens through a topK operation on the ranked scores, where the amount is related to the degree of reduction.

A concern is that tokens discarded in a layer might concentrate in a specific image area, potentially leading to spatial-centralized information loss.
Therefore, we develop a \textbf{local penalty strategy} that encourages a more uniform spatial distribution of discarded tokens.
Specifically, 
we represent the redundancy scoring vector $\mathbf{s}$ back to a 2D grid and partition it into non-overlapping
windows of size $W$, using padding for previously discarded tokens to maintain the 2D information.
Finally, we multiply the highest score within each window by a scaling coefficient, enhancing positive scores and diminishing negative ones.
This effectively suppresses the global prominence of other large scores within the windows, reducing their likelihood of being discarded.
As observed in the ablation study, 
this technique significantly enhances \textit{FiCoCo-V}.

\begin{figure}[!t]
  \centering
  \includegraphics[width=1\linewidth]{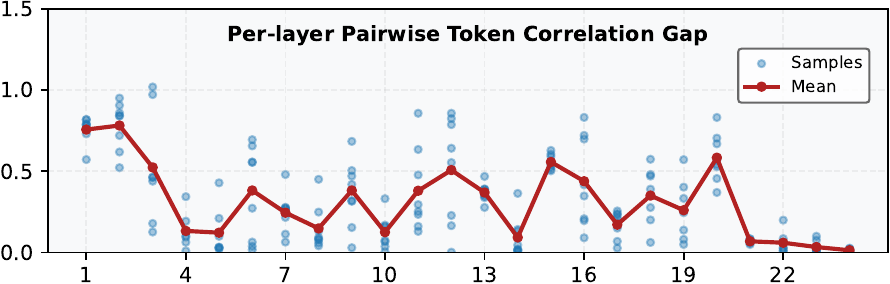}
  \caption{\textbf{Layer-wise distribution of Pairwise Token Correlation Gap.}
  We observe that the difference between the top-1 and top-2 correlation scores for each discarded token displays a high variance, highlighting the inadequacy of a fixed number of correlated preserved tokens.
  }  %
  \label{fig:many2many}
\end{figure}

\subsection{Correlation-based Information Recycling} 
\para{Correlate: Where should discarded information be recycled?}
We conduct a matrix that evaluates the correlation between each discarded token and all the preserved visual tokens.
Formally, given $N^{\mathbb{S}}$ discarded tokens, the matrix can be defined as $\mathbf{C} \in \mathbb{R}^{N^{\mathbb{S}} \times (N-N^{\mathbb{S}})}$.  %
For \textit{FiCoCo-V} in the vision encoder, attention 
weights inherently represent both the inter-token relationships and the flow of information, making them a measure of \textbf{direct} correlation.
Therefore, the correlation matrix can be conducted as $\mathbf{C}^v_{i,j} = \mathbf{A}^v_{i,j}.$

To select the preserved tokens that receive information from each discarded token, a topK operation can be applied on each row of the correlation matrix $\mathbf{C}$.
Here, $K$ is the number of the selected preserved tokens, where $K = 0$ is equivalent to token pruning and $K > 0$ is token merging (\textit{e.g.}, $K=1$ for ToMe~\cite{Bolya:ToMe}).
To find an appropriate $K$,
we measure the \textit{Pairwise Token Correlation Gap}—the difference between the top-1 and top-2 correlation scores—for each discarded token. 
Our \textit{FiCoCo-V} experiment computes this gap by discarding 8 tokens per layer across a 24-layer ViT.
In Figure~\ref{fig:many2many}, we observe that the distribution of correlation gaps varies significantly.
A few layers show a concentration of large gaps, indicating that discarded tokens can easily identify their single most correlated preserved token. 
In contrast, some layers have gaps concentrated near zero, suggesting each discarded token has multiple candidate preserved tokens with similarly high correlation values.
And more layers display distributions with high variance, highlighting the inadequacy of a fixed $K$ value.

According to the above analysis,
we devise a \textbf{token-adaptive} $K$.
Specifically, for the $i$-th discarded token, we compute the $\varepsilon$-th quantiles of the $i$-th row in the correlation matrix to determine a \textbf{token-wise threshold} $\tau_i$.
Then this threshold is re-applied to the row to identify the target tokens correlated to the $i$-th discarded token.
In other word, for the $j$-th preserved token, if $\mathbf{C}_{i,j} \geq \tau_i$, then this preserved token can be viewed as a correlated token for the $i$-th discarded token.
And the number of correlated tokens for each discarded token is dynamic and adaptive.
Therefore, we actually construct \textbf{``dense'' information pathways}, where the correlation matrix facilitates the tracking of the information propagation from each discarded token to the candidate tokens.
In contrast, a ``convergent'' correlation prompts all discarded tokens to merge into an additional token~\cite{Liang:EViT}.
Compared to that strategy,
our ``dense'' correlations spread discarded information more widely among the remaining tokens and empirically demonstrate better performance.

\begin{table*}[!t]
\tablestyle{5pt}{1.0}
\setlength\tabcolsep{0.5pt}
\def\w{20pt} 
\renewcommand{\arraystretch}{0.8}
\scalebox{0.95}{
\begin{tabular}{lcccccccccc}
\multicolumn{1}{p{5em}}{\centering \textbf{Method}} & 
\multicolumn{1}{p{5em}}{\centering \textbf{Source}} & 
\multicolumn{1}{p{5em}}{\centering \textbf{TFLOPs↓}} & 
\multicolumn{1}{p{5em}}{\centering \textbf{SQA}} & 
\multicolumn{1}{p{5em}}{\centering \textbf{VQA\textsuperscript{T}}} & 
\multicolumn{1}{p{5em}}{\centering \textbf{POPE}} & 
\multicolumn{1}{p{5em}}{\centering \textbf{GQA}} & 
\multicolumn{1}{p{5em}}{\centering \textbf{MMB}} & 
\multicolumn{1}{p{5em}}{\centering \textbf{VQAv2}} & 
\multicolumn{1}{p{5em}}{\centering \textbf{Avg}} & 
\multicolumn{1}{p{5em}}{\centering \textbf{Avg(\%)}} \\ 
\specialrule{1.5pt}{0pt}{0pt}

\multicolumn{11}{c}{\textit{TFLOPs=8.5}} \\
LLaVA-1.5-7B & \textit{NeurIPS23} & 8.5 & 69.5 & 58.2 & 86.4 & 62.5 & 66.1 & 79.1 & 70.3 & 100 \\

\midrule
\multicolumn{11}{c}{\footnotesize \textit{TFLOPs=3.3}$_{(\downarrow61.2\%)}$} \\
ToMe      & \textit{ICLR23}   & 3.3 & 65.2 & 52.1 & 72.4 & 54.3 & 60.5 & 68.0 & 62.1 & 88.3 \\
FastV     & \textit{ECCV24}   & 3.3 & 67.3 & 52.5 & 64.8 & 52.7 & 61.2 & 67.1 & 60.9 & 86.6 \\
SparseVLM & \textit{ICML25}   & 3.3 & 69.1 & 56.1 & 83.6 & 57.6 & 62.5 & 75.6 & 67.4 & 95.9 \\
PDrop     & \textit{CVPR25}   & 3.3 & 68.8 & 56.1 & 82.3 & 57.1 & 63.2 & 75.1 & 67.1 & 95.4 \\
PruMerge & \textit{ICCV25} & 3.3 & 67.9 & 54.3 & 71.3 & 54.3 & 59.6 & 70.6 & 63.0 & 89.6\\
\textit{\textbf{FiCoCo-V}} & Ours & 3.3 & 67.8 & 55.7 & 82.5 & 58.5 & 62.3 & 74.4 & 66.9 & 95.2 \\
\textit{\textbf{FiCoCo-L}} & Ours & 3.3 & \textbf{69.6} & \textbf{56.6} & \textbf{84.6} & \textbf{61.1} & \textbf{64.6} & \textbf{76.8} & \textbf{68.9} & \textbf{98.0} \\

\midrule
\multicolumn{11}{c}{\footnotesize \textit{TFLOPs=2.4}$_{(\downarrow71.8\%)}$} \\
ToMe      & \textit{ICLR23}   & 2.5 & 59.6 & 49.1 & 62.8 & 52.4 & 53.3 & 63.0 & 56.7 & 80.7 \\
FastV     & \textit{ECCV24}   & 2.5 & 60.2 & 50.6 & 59.6 & 49.6 & 56.1 & 61.8 & 56.3 & 80.1 \\
SparseVLM & \textit{ICML25}   & 2.5 & 67.1 & 54.9 & 80.5 & 56.0 & 60.0 & \textbf{73.8} & 65.4 & 93.0 \\
PDrop     & \textit{CVPR25}   & 2.5 & 68.3 & 55.1 & 82.3 & 56.0 & 61.1 & 72.9 & 65.9 & 93.8 \\
PruMerge & \textit{ICCV25} & 2.5 & 67.1 & 54.3 & 67.2 & 53.3 & 58.1 & 68.8 & 61.5 & 87.5 \\
\textit{\textbf{FiCoCo-V}} & Ours & 2.4 & 68.3 & 55.6 & 82.2 & 57.6 & 61.1 & 73.1 & 66.3 & 94.3 \\
\textit{\textbf{FiCoCo-L}} & Ours & 2.4 & \textbf{69.4} & \textbf{56.3} & \textbf{84.4} & \textbf{60.6} & \textbf{61.9} & 73.4 & \textbf{67.7} & \textbf{96.3} \\

\midrule
\multicolumn{11}{c}{\footnotesize \textit{TFLOPs=1.5}$_{(\downarrow82.4\%)}$} \\
ToMe      & \textit{ICLR23}   & 1.6 & 50.0 & 45.3 & 52.5 & 48.6 & 43.7 & 57.1 & 49.5 & 70.4 \\
FastV     & \textit{ECCV24}   & 1.6 & 51.1 & 47.8 & 48.0 & 46.1 & 48.0 & 61.8 & 50.5 & 71.8 \\
SparseVLM & \textit{ICML25}   & 1.5 & 62.2 & 51.8 & 75.1 & 52.4 & 56.2 & 68.2 & 61.0 & 86.8\\
PDrop     & \textit{CVPR25}   & 1.6 & 68.6 & 45.9 & 55.9 & 41.9 & 33.3 & 69.2 & 52.5 & 74.6 \\
PruMerge & \textit{ICCV25} & 1.5 & 68.1 & 54.0 & 65.3 & 51.9 & 55.3 & 67.4 & 60.3 & 85.8 \\
\textit{\textbf{FiCoCo-V}} & Ours & 1.5 & 68.4 & 55.5 & 79.8 & \textbf{54.9} & 60.2 & \textbf{72.1} & 65.2 & 92.7 \\
\textit{\textbf{FiCoCo-L}} & Ours & 1.5 & \textbf{69.5} & \textbf{55.7} & \textbf{82.1} & 53.2 & \textbf{61.5} & 69.7 & \textbf{65.3} & \textbf{92.8} \\

\shline
\multicolumn{11}{c}{\footnotesize \textit{TFLOPs=24.9}} \\
LLaVA-1.5-13B & \textit{NeurIPS23} & 24.9 & 71.4 & 61.3 & 86.2 & 63.4 & 68.0 & 80.0 & 71.7 & 100 \\
\midrule
\multicolumn{11}{c}{\footnotesize \textit{TFLOPs=15.4$_{(\downarrow47.6\%)}$}} \\
FastV & \textit{ECCV24} & 15.4 & 57.0 & 56.0 & 79.3 & 57.7 & 57.9 & - & 61.6 & 85.9 \\
SparseVLM & \textit{ICML25} & 15.4 & 69.9 & 49.9 & 81.1 & 57.9 & 65.8 & - & 64.9 & 90.5 \\
\textit{\textbf{FiCoCo-V}} & Ours & 15.4 & 72.1 & 57.2 & 82.3 & 59.2 & 63.1 & 76.8 & 68.5 & 95.5 \\
\textit{\textbf{FiCoCo-L}} & Ours & 15.4 & \textbf{72.4} & \textbf{58.3} & \textbf{83.1} & \textbf{60.1} & \textbf{65.2} & \textbf{77.6} & \textbf{69.5} & \textbf{96.9} \\

\specialrule{1.5pt}{0pt}{0pt}
\end{tabular}
}
\caption{
\textbf{Comparison results on LLaVA-1.5-7B/13B.} 
We evaluate \textit{FiCoCo} variants under various computational budgets, compared to baselines. 
Only shared datasets across both model sizes are included here.
}
\label{tab:main_results_combined}
\end{table*}

\para{Compress: How to effectively recycle information?}
After the correlate stage, each preserved token has a variable number of discarded tokens for updating itself.
A straightforward update strategy involves averaging each preserved token with all discarded tokens that correlated to it~\cite{Bolya:ToMe}.
However, as the number of discarded tokens increases, this strategy results in the preserved token having less information about itself after updates.
And excessive integration of information from discarded tokens into preserved tokens leads to performance degradation through progressive dilution of their original semantic content.
Therefore, our compression strategy must ensure the dominance of the preserved tokens.
Moreover, naive averaging results in the amount of information received by a preserved token being independent of its correlation to the discarded tokens.

According to the above discussion, we update the preserved tokens with a \textbf{self-preserving compression}.
Formally, we define the discarded tokens as a source set $\mathbb{S}$, and the preserved visual tokens as a target set $\mathbb{T}$.
Therefore, given the correlation matrix $\mathbf{C}$, we formulate the compression as
\begin{equation}
\begin{aligned}
\mathbf{X}^{\mathbb{T}}_j &\leftarrow \frac{\mathbf{X}^{\mathbb{T}}_j + \sum\limits_{i \in {\mathbb{I}_j}} \alpha_{ij}\mathbf{X}^{\mathbb{S}}_i }{1 + \sum\limits_{i \in {\mathbb{I}_j}}\alpha_{ij}}, \text{where} \ \mathbb{I}_j = \{i \in \mathbb{S} \text{ and } \mathbf{C}_{i,j} \geq \tau_i\}, \\
&\alpha_{ij} = \frac{\mathbf{C}_{i,j}}{\sum\limits_{j \in \mathbb{J}_i} \mathbf{C}_{i,j}}, \text{where} \ \mathbb{J}_i = \{j \in \mathbb{T} \text{ and } \mathbf{C}_{i,j} \geq \tau_i\},    \label{eq:ficoco-v-compress}
\end{aligned}
\end{equation}
where the weight $\alpha_{ij}$ quantifies the proportion of information from the $i$-th discarded token that is allocated to the $j$-th correlated token.
The strategy guarantees each preserved token preserves at least 50\% of its original information.
Moreover, the preserved token can receive more information from a discarded token with a strong correlation.

\section{Task-Aware Improvements for \textit{FiCoCo-L}}    \label{sec:ficoco-l}

Despite its promising performance, \textit{FiCoCo-V} identifies and removes visual tokens based solely on the visual content.
Applied within the vision encoder, such a task-agnostic method fails to preserve the essential visual information based on task context.
Therefore, we provide a task-aware solution \textit{\textbf{FiCoCo-L}} applied in the LLM decoder, leveraging textual priors to reduce visual token redundancy and recycle task-related visual information.
Specifically, \textit{FiCoCo-L} updates the redundancy calculation in the \textbf{\textit{filter}} stage, and the correlation calculation in the \textbf{\textit{correlate}} stage.

\para{Task-aware redundancy calculation.}
In the LLM decoder, since textual tokens directly encode task instructions, the attention weights that visual tokens received from textual tokens indicate their task relevance.
Therefore, we can calculate a \textbf{task-aware redundancy} as - $\frac{1}{M} \sum^{N+M}_{k=N+1} \mathbf{A}^l_{i, k}$, where $M$ denotes the number of textual tokens.
As a result, the overall redundancy for \textit{FiCoCo-L} is summarized as
{\setlength\abovedisplayskip{2mm}
\setlength\belowdisplayskip{2mm}
\begin{equation}
\mathbf{s}^l_i = \beta \frac{1}{N} \sum_{j=1}^N \mathbf{A}^l_{i,j} - (1 - \beta) \frac{1}{M} \sum^{N+M}_{k=N+1} \mathbf{A}^l_{i, k},
\end{equation}
}%
where the scalar hyperparameter $\beta$ balances the factors.

Empirically, we observe that the ``\textit{local penalty}'' strategy slightly degrades the performance of \textit{FiCoCo-L}.
We believe the reason is that this strategy weakens the task prior when encouraging spatial-uniform preservation of visual information.
Consequently, we remove the strategy in \textit{FiCoCo-L}.

\para{Task-aware correlation calculation.}
When calculating the correlation matrix for \textit{FiCoCo-L}, we explore an additional form of \textbf{indirect} semantic correlation, which leverages textual tokens as a bridge.
Specifically, when measuring the association between the $i$-th token and the $j$-th token, we sum the products of the attention weights from the $i$-th token to all textual tokens and from all textual tokens to the $j$-th token.
If the peak attention weights of the $i$-th token and the $j$-th token are concentrated on the same textual tokens, then the computed correlation between them is higher.
In summary, we have
\begin{equation}
\mathbf{C}^l_{i,j} = \gamma \mathbf{A}^l_{i,j} + (1 - \gamma) \frac{1}{M} \sum_{k = N+1}^{N+M}\mathbf{A}^l_{i,k} \cdot \mathbf{A}^l_{k,j},
\end{equation}
where $\gamma$ is the scalar hyperparameter for factor balance.

To facilitate a clearer understanding of the proposed methods, we provide a comprehensive and detailed explanation of our \textit{FiCoCo-V} and \textit{FiCoCo-L} processes in Appendix.
We also provide a theoretical estimation of the computing cost in Appendix.
Note that for clarity, our formula calculations are designed to target individual elements within vectors or matrices.
However, these operations can be efficiently tensorized in the practical implementation to facilitate batched inference.
And the implementation can be plug and play with less than 10 lines of additional code.
\section{Experiments}

\subsection{Experimental Setups}

We evaluate \textit{FiCoCo} on multiple MLLMs: LLaVA-1.5~\cite{Liu:LLaVA-1.5}, LLaVA-NeXT~\cite{Liu:LLaVA-NeXT}, and Qwen2-VL~\cite{Wang:Qwen2-VL} for image understanding, and Video-LLaVA~\cite{Lin:Video-LLaVA} for video understanding. \textit{FiCoCo} is benchmarked against mainstream token reduction methods: ToMe~\cite{Bolya:ToMe}, FastV~\cite{Chen:FastV}, SparseVLM~\cite{Zhang:SparseVLM}, PDrop~\cite{Xing:PyramidDrop}, and PruMerge~\cite{Shang:LLaVA-PruMerge}.
Configurations and benchmark details are provided in Appendix.

\subsection{Main Comparisons}

\para{Results on LLaVA-1.5-7B/13B.}
Table~\ref{tab:main_results_combined} presents the performance of \textit{FiCoCo} across benchmarks based on LLaVA-1.5-7B/13B. The LLaVA-1.5-7B results yield two key findings: \textbf{(1)} Both \textit{FiCoCo-V} and \textit{FiCoCo-L} consistently outperform existing training-free methods across different computational budgets. Under extreme compression (TFLOPs = 1.5, ~10\% visual tokens), both variants achieve \textbf{>92\%} average accuracy, surpassing the second-best SparseVLM by approximately \textbf{6\%}, demonstrating the effectiveness of our information recovery mechanism. \textbf{(2)} \textit{FiCoCo-L} outperforms \textit{FiCoCo-V} when computational budgets are generous, as it captures task-relevant visual tokens and maximally \textbf{focuses} on question-related regions. However, under constrained budgets, both variants achieve similar performance (with an average difference of only \textbf{0.1\%}). This convergence arises because, under severe token constraints, the distinction between preserving visual saliency (\textit{FiCoCo-V}) and task-specific relevance (\textit{FiCoCo-L}) diminishes—both are compelled to retain only the most essential visual elements for maintaining core model functionality. Moreover, under the LLaVA-1.5-13B setting, \textit{FiCoCo-V/L} achieves superior performance with only $\sim$22\% visual tokens (15.4 TFLOPs), outperforming the strongest baseline SparseVLM by \textbf{5.0\%} and \textbf{6.4\%} in average accuracy. Additional results in Tables~\ref{7B_more_results} further validate this advantage.

\para{Results on LLaVA-NeXT-7B.}
We impose two computational constraints: TFLOPs are set to 5.0 for PDrop to match reported results, and 2.9 for SparseVLM and \textit{FiCoCo}. Table~\ref{tab:llava_next} shows that, under TFLOPs = 2.9, \textit{FiCoCo-V} and \textit{FiCoCo-L} outperform SparseVLM by \textbf{3.6\%} and \textbf{4.2\%}, respectively. Moreover, despite operating under a lower TFLOPs budget than PDrop, our methods \textit{FiCoCo-V} and \textit{FiCoCo-L} consistently outperform PDrop by \textbf{1.9\%} in terms of average accuracy, highlighting their superior efficiency and robustness in handling dense visual token scenarios within resource-constrained settings.

\begin{table}[!t]
\tablestyle{3pt}{1.0}
\setlength\tabcolsep{2.5pt}
\def\w{20pt} 
\renewcommand{\arraystretch}{0.8}
\scalebox{0.97}{ 
    \begin{tabular}{l cccc cc} 
    \textbf{Method} & \textbf{MMB} & \textbf{SQA} & \textbf{VQA\textsuperscript{T}} & \textbf{MMMU} & \textbf{Avg} & \textbf{Avg (\%)} \\
    \shline
    \multicolumn{7}{c}{\footnotesize \textit{TFLOPs=42.7}} \\
    LLaVA-NeXT-7B & 67.9 & 70.2 & 61.3 & 35.1 & 58.6 & 100.0 \\
    \hline
    \multicolumn{7}{c}{\footnotesize \textit{TFLOPs=5.0}$_{(\downarrow88.3\%)}$} \\
    PDrop & 63.4 & 67.5 & 54.4 & 29.8 & 53.8 & 91.7 \\
    \hline
    \multicolumn{7}{c}{\footnotesize \textit{TFLOPs=2.9}$_{(\downarrow93.2\%)}$} \\
    SparseVLM & 63.1 & 67.5 & 46.3 & 32.8 & 52.4 & 89.4 \\
    \textbf{\textit{FiCoCo-V}} & 60.5 & \textbf{68.1} & \textbf{55.3} & 34.1 & 54.5 & 93.0 \\
    \textbf{\textit{FiCoCo-L}} & \textbf{63.6} & 67.9 & 53.1 & \textbf{34.8} & \textbf{54.9} & \textbf{93.6} \\
    \shline
    \end{tabular}%
}
\caption{\textbf{Comparison with LLaVA-NeXT-7B on cross-image understanding benchmarks.}}
\label{tab:llava_next} %
\end{table}

\begin{table}[!t]
\tablestyle{3pt}{1.0} %
\setlength\tabcolsep{3pt}
\renewcommand{\arraystretch}{0.8}
\scalebox{0.97}{ 
\begin{tabular}{c l c c c c c}
\textbf{Token} & \textbf{Method} & \textbf{MMB} & \textbf{POPE} & \textbf{VQA\textsuperscript{T}} & \textbf{Avg} & \textbf{Avg (\%)} \\
\shline
\multicolumn{7}{c}{\footnotesize \textit{Base.} $\approx$1300} \\
 & Qwen2-VL-7B & 80.5 & 86.4 & 84.3 & 83.7 & 100.00 \\
\hline
\multirow{3}{*}{600} 
 & SparseVLM & 79.6 & 86.5 & 80.3 & 82.1 & 98.09 \\
 & \textit{FiCoCo-V} & 79.9 & 86.5 & 81.2 & 82.5 & 98.57 \\
 & \textit{FiCoCo-L} & 80.1 & 86.3 & 81.4 & \textbf{82.6} & \textbf{98.69} \\
\hline
\multirow{3}{*}{500} 
 & SparseVLM & 78.8 & 86.3 & 79.0 & 81.4 & 97.25 \\
 & \textit{FiCoCo-V} & 79.1 & 86.6 & 79.8 & \textbf{81.8} & \textbf{97.73} \\
 & \textit{FiCoCo-L} & 78.9 & 86.1 & 79.7 & 81.6 & 97.49 \\
\hline
\multirow{3}{*}{400} 
 & SparseVLM & 79.0 & 85.8 & 77.1 & 80.6 & 96.29 \\
 & \textit{FiCoCo-V} & 79.1 & 86.0 & 78.3 & \textbf{81.1} & \textbf{96.89} \\
 & \textit{FiCoCo-L} & 78.6 & 85.9 & 78.3 & 80.9 & 96.66 \\
\shline
\end{tabular}
}
\caption{\textbf{Comparison with Qwen2-VL-7B under different token budgets.} \textit{FiCoCo} is compared against SparseVLM under 600/500/400 tokens on MMB, POPE, and VQA\textsuperscript{T}.}
\label{tab:token_compression_comparison}
\end{table}

\para{Results on Qwen2-VL.}
Following SparseVLM’s settings on Qwen2-VL, we compress about 54.5\% of visual tokens. 
Since Qwen2-VL lacks a \texttt{[CLS]} token, an equivalent token averaging scheme is used in \textit{FiCoCo-V}. 
As shown in Table~\ref{tab:token_compression_comparison}, under this ratio, \textit{FiCoCo} maintains over \textbf{98\%} accuracy and surpasses SparseVLM. 
Further, compressing each additional 100 tokens yields only a 0.8\% drop, highlighting \textit{FiCoCo}’s robustness and the effectiveness of the averaging scheme.

\para{Results on Video-LLaVA.}
For fair evaluation, the number of video tokens is limited to 136 (about 6.6\% of all visual tokens). 
As shown in Table~\ref{tab:video_llava_comparison}, both \textit{FiCoCo-V} and \textit{FiCoCo-L} reach over \textbf{90\%} of Video-LLaVA’s performance, 
while \textit{FiCoCo-V} surpasses FastV and SparseVLM by \textbf{40.7\%} and \textbf{6.3\%}, respectively, demonstrating its superiority in video understanding. 
The stronger performance of \textit{FiCoCo-V} over \textit{FiCoCo-L} likely stems from higher visual redundancy in videos, where compression aids attention to salient content.

\begin{table}[!t]
\tablestyle{3pt}{1.0}
\setlength\tabcolsep{1pt}
\def\w{20pt} 
\renewcommand{\arraystretch}{0.8}
\scalebox{0.97}{
    \begin{tabular}{l cccc cc}
    \textbf{Method} & \textbf{TGIF} & \textbf{MSVD} & \textbf{MSRVTT} & \textbf{ActivityNet} & \textbf{Avg} & \textbf{Avg(\%)} \\
    \shline
    \multicolumn{7}{c}{\footnotesize \textit{TFLOPs=29.7}} \\
    Video-LLaVA & 47.1 & 69.8 & 56.7 & 43.1 & 54.2 & 100.0 \\
    \hline
    \multicolumn{7}{c}{\footnotesize \textit{TFLOPs=2.6}$_{(\downarrow91.2\%)}$} \\
    FastV & 23.1 & 38.0 & 19.3 & 30.6 & 27.8 & 52.1 \\
    SparseVLM & 44.7 & \textbf{68.2} & 31.0 & 42.6 & 46.9 & 86.5 \\
    \textbf{\textit{FiCoCo-V}} & \textbf{43.1} & 67.4 & 47.8 & \textbf{42.8} & \textbf{50.3} & \textbf{92.8} \\
    \textbf{\textit{FiCoCo-L}} & 44.3 & 64.5 & \textbf{49.2} & 40.1 & 49.5 & 91.4 \\
    \shline
    \end{tabular}%
}
\caption{
\textbf{Comparison results on video understanding benchmarks with Video-LLaVA.}
}
\label{tab:video_llava_comparison}
\end{table}

\begin{table}[!t]
  \tablestyle{5pt}{1.0}
\setlength\tabcolsep{2.3pt}
\def\w{20pt} 

\scalebox{1}{
    \begin{tabular}{l|lcc}
    \multirow{2}[1]{*}{\textbf{Stage}} & \textbf{Method} & \textbf{SQA} & \textbf{TextVQA} \\
          & \textit{\textbf{FiCoCo-V}} & \textbf{68.37} & \textbf{55.46} \\
    \shline
    \multirow{3}[2]{*}{Filter} & w/o vision-aware redundancy & 67.81 & 52.51 \\
          & w/o semantic-aware redundancy & 64.67 & 48.74 \\
          & w/o local penalty & 68.12 & 53.24 \\
    \hline
    \multirow{4}[2]{*}{Correlate} & fixed K=0 & 67.82 & 53.56 \\
          & fixed K=1 & 67.43 & 46.97 \\
          & fixed K=2 & 67.21 & 51.36 \\
          & convergent correlation & 67.60 & 54.38 \\
    \hline
    \multirow{1}[1]{*}{Compress} 
          & average compression & 67.92 & 53.34 \\
          
    \end{tabular}%
    }
  \caption{
  \textbf{Ablation results of \textit{FiCoCo-V}.}
  }
  \label{tab:FiCoCo-V ablation study}
\end{table}%
\subsection{Ablation Study}    \label{sec:ablation}

To further validate the effectiveness of our design at each stage, we conduct extensive ablation studies on the SQA and TextVQA benchmarks under a fixed computational budget of 1.5 TFLOPs.
In Table~\ref{tab:FiCoCo-V ablation study}, we ablate all three stages of \textbf{\textit{FiCoCo-V}} to analyze their individual contributions.

\noindent• \textbf{Filter.} 
Both vision-aware and semantic-aware redundancy improve the identification of discarded tokens.
Notably, semantic-aware redundancy has a more significant impact on the final performance.
This indicates that token reduction within the vision encoder should prioritize the retention of tokens rich in global semantic information.
Additionally, we observe that by promoting a spatially uniform distribution of discarded tokens, the local penalty strategy aids in preserving visual information.

\noindent• \textbf{Correlate.} We evaluate fixed $K$ values of 0 (pruning, \textit{i.e.}, no correlation-based recycling), 
1 (single-token recycling), and 2 (multi-token recycling). 
While the token-adaptive $K$ strategy performs best, an intriguing result is that $K{=}0$ surpasses the other two. 
This likely occurs because small fixed $K$ values limit information sources for correlated tokens, causing over-dilution and noise. 
Thus, pruning yields better performance. 
Moreover, our ``dense'' correlation outperforms the ``convergent'' variant, which compresses discarded tokens into one; 
retrieving information while preserving token integrity proves more effective.

\noindent• \textbf{Compress.}
Our self-preserving compression outperforms directly averaging the features, indicating that the calculated weights can effectively regulate the contribution of information sources in the updates of correlated tokens.

In Table~\ref{tab:FiCoCo-L ablation study}, we ablate all three stages for \textbf{\textit{FiCoCo-L}}:

\begin{table}[!t]
  \tablestyle{5pt}{1.0}
\setlength\tabcolsep{2pt}
\def\w{20pt} 
\scalebox{1}{
    \begin{tabular}{l|lcc}
    \multirow{2}[1]{*}{\textbf{Stage}} & \textbf{Method} & \textbf{SQA} & \textbf{TextVQA} \\
          & \textit{\textbf{FiCoCo-L}} & \textbf{69.46} & \textbf{55.72} \\
    \shline
    \multirow{3}[2]{*}{Filter} & w/o vision-aware redundancy & 69.16 & 55.43 \\
          & w/o task-aware redundancy & 68.22 & 55.64 \\
          & w/ local penalty & 68.79 & 55.38 \\
    \hline
    \multirow{6}[2]{*}{Correlate} & w/o indirect correlation & 68.89 & 54.78 \\
          & w/o direct correlation & 68.45 & 55.45 \\
          & fixed K=0 & 68.96 & 50.33 \\
          & fixed K=1 & 68.57 & 50.11 \\
          & fixed K=2 & 68.32 & 50.18 \\
          & convergent correlation & 67.80 & 54.89 \\
    \hline
    \multirow{1}[1]{*}{Compress} 
          & average compression & 68.32 & 54.66 \\
    \end{tabular}%
    }
  \caption{
   \textbf{Ablation results of \textit{FiCoCo-L}.}
  }
\label{tab:FiCoCo-L ablation study}
\end{table}%

\noindent• \textbf{Filter.}
Although both vision-aware and task-aware redundancies contribute to redundancy estimation, neither dominates. This may be because the attention mechanism in LLMs captures stable token dependencies, reducing the need for redundancy measurement to rely heavily on semantic cues. Moreover, applying the local penalty strategy in \textit{FiCoCo-L} slightly degrades performance, likely because enforcing spatial uniformity of token retention disrupts the redundancy assessments already established by attention.

\noindent• \textbf{Correlate.}
It is observed that both correlations enhance the identification of relevant tokens, improving performance across datasets. Similar to \textit{FiCoCo-V}, adopting a token-adaptive $K$ with ``dense'' correlations proves optimal.

\noindent• \textbf{Compress.}
Updating the preserved tokens with a self-preserving compression still achieves better performance.

\subsection{Qualitative Analysis}
\begin{figure}[!t]
  \centering
  \includegraphics[width=\linewidth]{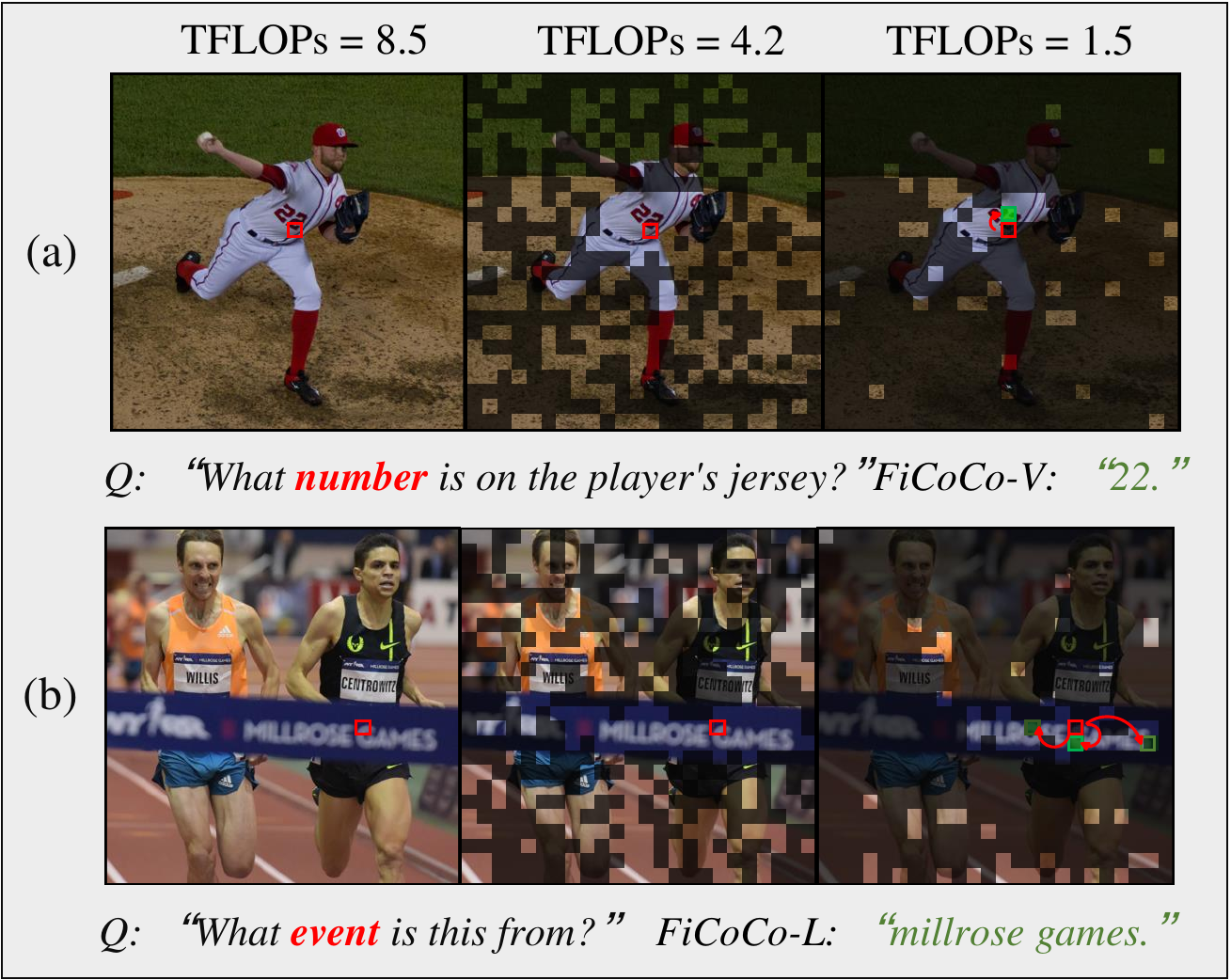}
  \caption{
  \textbf{Visualizations of token discard and information recycling by (a) \textit{FiCoCo-V} and (b) \textit{FiCoCo-L}.} Red: traced patch token; Green: recycling destination.
  }
  
  \label{fig:token compress}
\end{figure}

We visualize the discarded tokens of \textit{FiCoCo-V} (Figure~\ref{fig:token compress}a) and \textit{FiCoCo-L} (Figure~\ref{fig:token compress}b) under varying compression levels across VQA scenarios. 
Tokens highly relevant to answers are highlighted with red boxes to assess information preservation. 
A token linked to ``\textit{2}'' is traced in Figure~\ref{fig:token compress}a, and one linked to ``\textit{GAMES}'' in Figure~\ref{fig:token compress}b. 
In both cases, as compression increases (TFLOPs from 4.2 to 1.5), more tokens—including key ones—are discarded, reducing critical information. We trace information recycling from discarded tokens (red arrows) and highlight correlated tokens (green boxes), 
where transparency reflects retained information. 
These correlated tokens aggregate essential cues and support question answering. 
Notably, discarded information is distributed across multiple correlated tokens, 
enhancing comprehension of salient regions (Figure~\ref{fig:token compress}b), 
qualitatively validating our method.

\subsection{Efficiency Analysis}
As shown in Table~\ref{tab:grouped_tflops_throughput}, we present the trends of throughput and TFLOPs changes after applying \textit{FiCoCo} in the LLaVA-NeXT and LLaVA-1.5 architectures. Introducing \textit{FiCoCo} into LLaVA-NeXT reduces TFLOPs by 93.2\%, increasing throughput by $2.08\times$ (\textit{FiCoCo-V}) and $1.71\times$ (\textit{FiCoCo-L}). In LLaVA, a TFLOPs reduction of 82.4\% yields throughput gains of $1.43\times$ and $1.29\times$, respectively. These results highlight \textit{FiCoCo}’s ability to substantially lower computational overhead while enhancing throughput.

\begin{table}[!t]
\tablestyle{3pt}{1.0}
\setlength\tabcolsep{2pt}

\scalebox{0.97}{
    \begin{tabular}{l cc cc}
        \multirow{2}{*}{\textbf{Method}} & \multicolumn{2}{c}{\textbf{LLaVA-NeXT-7B}} & \multicolumn{2}{c}{\textbf{LLaVA-1.5-7B}} \\
        & \textbf{TFLOPs} & \textbf{Throughput} & \textbf{TFLOPs} & \textbf{Throughput} \\
        \shline
        Vanilla & 42.7 & 3.8 & 8.5 & 8.99 \\
        \textit{FiCoCo-V}   & 2.9 {\scriptsize(\textbf{\textdownarrow~93.2\%})} & 7.9 {\scriptsize(\textbf{\textuparrow~107.9\%})} 
                   & 1.5 {\scriptsize(\textbf{\textdownarrow~82.4\%})} & 12.9 {\scriptsize(\textbf{\textuparrow~43.5\%})} \\
        \textit{FiCoCo-L}   & 2.9 {\scriptsize(\textbf{\textdownarrow~93.2\%})} & 6.5 {\scriptsize(\textbf{\textuparrow~71.1\%})} 
                   & 1.5 {\scriptsize(\textbf{\textdownarrow~82.4\%})} & 11.6 {\scriptsize(\textbf{\textuparrow~29.1\%})} \\
        \shline
    \end{tabular}
}
\caption{\textbf{Efficiency analysis of \textit{FiCoCo}.} Lower TFLOPs and higher throughput (img/s) indicate better efficiency.}
\label{tab:grouped_tflops_throughput}
\end{table}

\section{Conclusion}

In this paper, we propose a ``\textit{filter-correlate-compress}'' acceleration framework to systematically eliminate visual redundancy in MLLMs through a principled three-stage pipeline. The \textit{filter} stage performs dynamic redundancy-based token discard using variation-aware thresholds, while the \textit{correlate} stage identifies semantic relationships between tokens and the \textit{compress} stage jointly enables correlation-based information recycling, thereby significantly reducing computational complexity while preserving critical multimodal information. The effectiveness of our framework is rigorously demonstrated through specialized variants for vision encoders (\textit{FiCoCo-V}) and LLM decoders (\textit{FiCoCo-L}), achieving consistent acceleration benefits across diverse MLLM architectures for both image and video understanding tasks with minimal accuracy degradation.

\clearpage
\section{Acknowledgments}
This work was supported in part by the National Natural Science Foundation of China (No. 62301432), the Natural Science Basic Research Program of Shaanxi Province (Grant No. QCYRCXM-2023-057), the Fundamental Research Funds for the Central Universities, and the Guangdong Basic and Applied Basic Research Foundation (Grant No. 2025A1515011119). It was also supported by the Chengdu Science and Technology Program (Grant No. 2025-YF12-00006-RC), the Police Integration Computing Key Laboratory of Sichuan Province (Grant No. JWRH202502002), and the Open Fund of the Key Laboratory of the Ministry of Education on Artificial Intelligence in Equipment (Grant No. 2024-AAIE-KF04-03). This work was supported by the National Science and Technology Innovation 2030 - Major Project (Grant No. 2022ZD0208800), and NSFC General Program (Grant No. 62176215)
\bibliography{ref}

\clearpage
\clearpage
\setcounter{page}{1}

\section*{Appendix}

In the appendix, we provide the theoretical calculation of FLOPs, detailed implementation information, extended experiments and analyses, and a comprehensive explanation of our proposed methods.

\section{Theoretical FLOPs Calculation}
\label{sec:theoretical_FLOPs}
Here we consider a hypothetical scenario to analyze the changes in FLOPs before and after applying \textit{FiCoCo-V} and \textit{FiCoCo-L}. In this context, the hidden State dimension in a single transformer layer is denoted as \( D \), while the feed-forward layer dimension is represented by \( H \). The total number of visual tokens is represented by \( N \), with \( N^{\mathbb{S}} \) denoting the number of compressed visual tokens per layer.

Additionally, \( M \) represents the number of text tokens. To simplify the equations, we define:
\[
N' = N - N^{\mathbb{S}}, \quad P = N + M, \quad P' = N' + M.
\]
Here, \( P \) represents the total number of visual and text tokens before compression, while \( P' \) represents the total tokens after compression. Finally, for \textit{FiCoCo-V}, we have:
\begin{equation}
\begin{aligned}
\text{FLOPs}_{\text{before}} &= 4ND^2 + 2N^2D + 2NDH, \\
\text{FLOPs}_{\text{after}}  = &4N'D^2 + 2(N')^2D + 2N'DH, \\
\Delta         = 4N^{\mathbb{S}}D^2& + 2\left(NN^{\mathbb{S}} - (N^{\mathbb{S}})^2\right)D + 2N^{\mathbb{S}}DH.
\end{aligned}
\end{equation}
For \textit{FiCoCo-L}, we have:
\begin{equation}
\begin{aligned}
\text{FLOPs}_{\text{before}} &= 4PD^2 + 2P^2D + 2PDH, \\
\text{FLOPs}_{\text{after}} = &4P'D^2 + 2(P')^2D + 2P'DH, \\
\Delta = 4N^{\mathbb{S}}D^2& + 2\left(2NN^{\mathbb{S}} - (N^{\mathbb{S}})^2\right)D + 2N^{\mathbb{S}}DH.
\end{aligned}
\end{equation}

We now analyze the additional FLOPs introduced by the internal operations of \textit{FiCoCo-V} and \textit{FiCoCo-L}. As described in Sec.~\ref{method}, the primary computational overhead for \textit{FiCoCo-V} stems from the redundancy score calculation, the determination of token-adaptive K values, and the token updating process. In comparison, \textit{FiCoCo-L} incorporates similar steps but introduces an additional interaction with the indirect text matrix during the correlate phase, resulting in a higher computational complexity. The variable \( N^{\mathbb{T}} \) represents the number of target tokens. However, since both \textit{FiCoCo-V} and \textit{FiCoCo-L} only operate on visual tokens, their FLOPs calculations are nearly identical.
For \textit{FiCoCo-V}, we have: \begin{equation} \text{FLOPs}= N^2 + 2N + N^{\mathbb{S}} (N^{\mathbb{T}} + 2D + 1) + D. \end{equation}
For \textit{FiCoCo-L}, we have: \begin{equation} \text{FLOPs}= 2(N^2 + 2N) + N^{\mathbb{S}} (N^{\mathbb{T}} + 2D + 1) + D. \end{equation}

Based on the above analysis, the additional FLOPs introduced by \textit{FiCoCo-V} and \textit{FiCoCo-L} are negligible compared to the significant reduction in FLOPs ( $\Delta$ ) achieved through token compression. Specifically, while $\Delta$ grows quadratically with the hidden State dimension $D$, the additional FLOPs primarily grow linearly, making their impact inconsequential in practical scenarios.

\section{Implementation Details}
\label{sec:more_imple}
\noindent \textbf{Experimental Setup Details.}  
For \textit{FiCoCo}, we adopt the LLaVA-1.5-7B/13B models~\cite{Liu:LLaVA-1.5} and employ the following settings: (1) $\lambda=0.35$ in filter stage of \textit{FiCoCo-V}, (2) $\beta=0.6$ in filter stage of \textit{FiCoCo-L}, (3) $\gamma=0.6$ in correlate stage of \textit{FiCoCo-L}, (4) scaling coefficient$=$2 in local penalty strategy, (5) $\varepsilon=0.998$ to determine the token-wise threshold in compress stage.
We provide sensitivity analyses of these hyperparameters in the latter half of this appendix.  %
For the local penalty strategy, we fix a $2\times2$ window across all layers.
Since the effectiveness of our \textit{FiCoCo} is based on the reliability of attention mechanisms,
we delay the token reduction until the attention converges to stability.
Specifically, in \textit{FiCoCo-V}, the token compression starts at the 12-th layer of the vision encoder, while in \textit{FiCoCo-L}, it starts at the 4-th layer of the LLM.
All experiments are conducted on a single A800 80GB GPU.

\noindent \textbf{Backbone Details.}  We detail the involved backbones as follows.
\begin{itemize} 
\item \textbf{LLaVA-1.5}~\cite{Liu:LLaVA-1.5}: An improved vision-language model built on the LLaVA framework that integrates high-resolution image processing and enhanced vision-language alignment. It employs CLIP-ViT-L/14 as the vision encoder and a Vicuna-based language backbone, with an optimized projection layer for visual token mapping. LLaVA-1.5 supports resolution up to 448×448 and achieves stronger accuracy across multiple benchmarks while maintaining efficient inference.
\item \textbf{LLaVA-NeXT}~\cite{Liu:LLaVA-NeXT}: A strong vision-language baseline that adopts the AnyRes strategy to preserve fine-grained image details by quadrupling input resolution. This leads to a substantial increase in the number of visual tokens and significantly higher computational cost.
\item \textbf{Qwen2-VL}~\cite{Wang:Qwen2-VL}: A multimodal model that introduces a cross-attention-based vision-language adapter to compress image features into 256 tokens using learnable query embeddings. To preserve spatial details, 2D absolute positional encodings are incorporated into the attention mechanism. The compressed features are then fed into the Qwen language model for downstream tasks.
\item \textbf{Video-LLaVA}~\cite{Lin:Video-LLaVA}: A video-based extension of LLaVA that utilizes Language-Bind as the visual encoder to process 8-frame clips, with 256 video tokens per frame. Following its original protocol, we adopt ChatGPT-based scoring as the main evaluation metric. 
\end{itemize}

\noindent \textbf{Benchmark Details.}  
We assess the performance of \textit{FiCoCo} across a comprehensive suite of multimodal understanding benchmarks, spanning both image-based and video-based reasoning tasks:

\begin{itemize}
    \item \textbf{GQA}~\cite{Hudson:GQA}: Comprises 113,018 real-world images annotated with scene graphs to support structured visual reasoning.
    
    \item \textbf{VizWiz}~\cite{Gurari:VizWiz}: Contains over 39,000 questions posed by blind users, accompanied by low-quality images and characterized by conversational query patterns.
    
    \item \textbf{SQA}~\cite{Lu:ScienceQA}: Encompasses 21,000 scientific questions covering 26 distinct domains and a wide range of 379 reasoning skills.
    
    \item \textbf{VQA-T}~\cite{Singh:TextVQA}: Involves 28,000 high-resolution images aimed at evaluating the understanding of textual content embedded within complex visual scenes.

    \item \textbf{POPE}~\cite{Li:POPE}: Offers 14,000 image-question pairs focusing on binary judgments for identifying object hallucinations.
    
    \item \textbf{MME}~\cite{Fu:MME}: Provides a collection of 15,000 images encompassing 14 distinct subdomains of perceptual and cognitive reasoning.
    
    \item \textbf{MMB}~\cite{Liu:MMBench}: Presents 20,000 multiple-choice questions that assess robust visual reasoning across various categories.
    
    \item \textbf{MMB-CN}~\cite{Liu:MMBench}: Serves as the Chinese counterpart of MMB, consisting of 20,000 questions tailored for cross-lingual evaluation.
    
    \item \textbf{MM-Vet}~\cite{Yu:MM-Vet}: Comprises 16,000 high-resolution questions spanning visual recognition, OCR, factual knowledge, and spatial reasoning.
    \item \textbf{TGIF}~\cite{Jang:TGIF}: A collection of animated GIFs paired with natural language descriptions. It focuses on short, dynamic visual content with fine-grained motion, and is widely used for video-to-text tasks such as captioning and retrieval.

    \item \textbf{MSVD}~\cite{Xu:MSVD}: The Microsoft Research Video Description dataset consists of around 2,000 short YouTube videos, each annotated with multiple human-written captions. It supports multilingual evaluation and is commonly used for video captioning.

    \item \textbf{MSRVTT}~\cite{Xu:MSVD}: This dataset contains 10,000 video clips from 20 diverse categories, each accompanied by 20 captions. It serves as a standard benchmark for general-purpose video understanding and text-video alignment tasks.

    \item \textbf{ActivityNet}~\cite{Yu:ActNet}: Built on top of the ActivityNet dataset, it features untrimmed videos annotated with temporally localized natural language descriptions. It is designed for dense video captioning and temporal event localization.

\end{itemize}

\section{More Experiments and Analysis}
\label{sec:more_exp}
\begin{table}[!t]
\centering
\small  %
\setlength\tabcolsep{0.5pt}  %
\renewcommand{\arraystretch}{1.2} %
\begin{tabular}{p{4em}ccccc}  %
    \multicolumn{1}{c}{\textbf{Method}} & 
    \textbf{TFLOPs↓} & 
    \textbf{Vizwiz} & 
    \textbf{MM-Vet} & 
    \textbf{MMBCN} & 
    \textbf{LLaVA-W} \\
    \specialrule{1.5pt}{0pt}{0pt}
    
    \multicolumn{6}{c}{\footnotesize \textit{TFLOPs = 8.5}} \\
    \multirow{2}[1]{*}{LLaVA-1.5} & \multirow{2}[1]{*}{8.5} & 50 & 31.6 & 59.3 & 63.7 \\
    &       & 100.0\% & 100.0\% & 100.0\% & 100.0\% \\

    \midrule
    \multicolumn{6}{c}{\footnotesize \textit{TFLOPs=3.3}$_{(\downarrow61.2\%)}$} \\
    \multirow{2}[1]{*}{\textit{FiCoCo-V}} & \multirow{2}[1]{*}{3.3} & \textbf{51.5} & 29.7 & \textbf{55.3} & \textbf{60.4} \\
    &       & \textbf{103.0\%} & 94.0\% & \textbf{93.3\%} & \textbf{94.8\%} \\
    \multirow{2}[0]{*}{\textit{FiCoCo-L}} & \multirow{2}[0]{*}{3.3} & 48.7 & \textbf{31.4} & 53.6 & 60.3 \\
    &       & 97.4\% & \textbf{99.4\%} & 90.4\% & 94.7\% \\

    \midrule
    \multicolumn{6}{c}{\footnotesize \textit{TFLOPs=2.4}$_{(\downarrow71.8\%)}$} \\
    \multirow{2}[1]{*}{\textit{FiCoCo-V}} & \multirow{2}[1]{*}{2.4} & \textbf{49.4} & 28.2 & \textbf{54.3} & 56.6 \\
    &       & \textbf{98.8\%} & 89.2\% & \textbf{91.6\%} & 88.9\% \\
    \multirow{2}[0]{*}{\textit{FiCoCo-L}} & \multirow{2}[0]{*}{2.4} & 48.4 & \textbf{30.1} & 53.5 & \textbf{59.4} \\
    &       & 96.8\% & \textbf{95.3\%} & 90.2\% & \textbf{93.3\%} \\

    \midrule
    \multicolumn{6}{c}{\footnotesize \textit{TFLOPs=1.5}$_{(\downarrow82.4\%)}$} \\
    \multirow{2}[1]{*}{\textit{FiCoCo-V}} & \multirow{2}[1]{*}{1.5} & \textbf{52.4} & 26.8 & 53.0 & \textbf{58.6} \\
    &       & \textbf{104.8\%} & 84.8\% & 89.4\% & \textbf{92.0\%} \\
    \multirow{2}[0]{*}{\textit{FiCoCo-L}} & \multirow{2}[0]{*}{1.5} & 48.2 & \textbf{27.4} & \textbf{53.3} & 57.3 \\
    &       & 96.4\% & \textbf{86.7\%} & \textbf{89.9\%} & 90.0\% \\
    \shline
\end{tabular}
\caption{\textbf{Additional results of \textit{FiCoCo} on LLaVA-1.5-7B.}}
\label{7B_more_results}
\vspace{-3mm}
\end{table}

\subsection{More Experiments on LLaVA-1.5-7B}

In Figure~\ref{7B_more_results}, more comparison results further presents the performance of our method on VizWiz~\cite{Gurari:VizWiz}, MM-Vet~\cite{Yu:MM-Vet}, MMBCN~\cite{Liu:MMBench}, and LLAVA-W\cite{Liu:LLaVA}. The results indicate that even with an 82.4\% reduction in TFLOPs, both \textit{FiCoCo-V} and \textit{FiCoCo-L} maintain an average accuracy exceeding 90\%, effectively preserving the capabilities of the MLLM.

\subsection{Disscussion about Evaluation without {\texttt{[CLS]}} token} \label{sec:without_CLS}

\begin{figure}[!t]
  \centering
   \includegraphics[width=\linewidth]{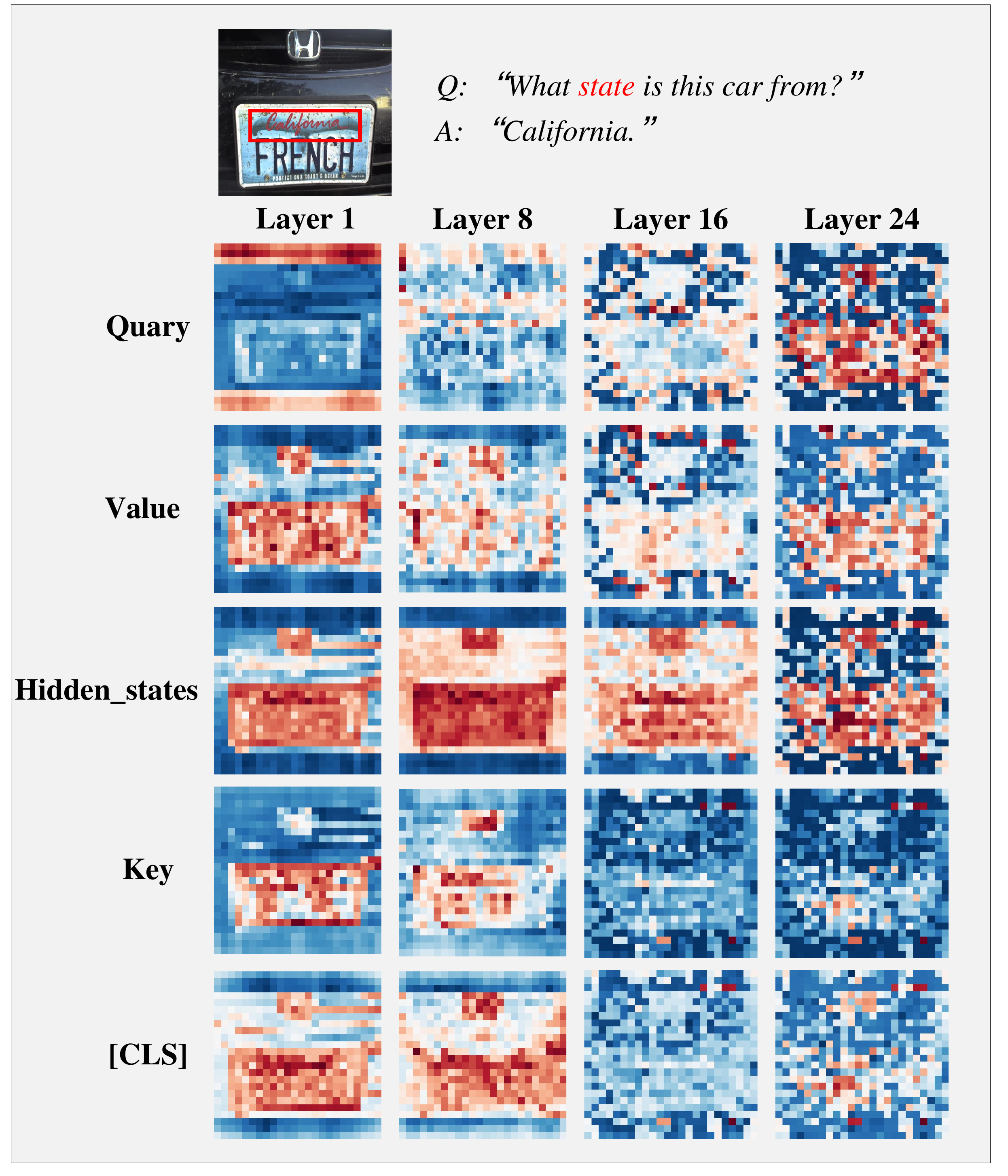}
   \caption{\textbf{Heatmap visualization using different inputs as the mean reference.}}  %
   \label{fig:ana_qkv}
   \vspace{-4mm}
\end{figure}

A key component of our \textit{FiCoCo-V}'s semantic-aware redundancy metric relies on the \texttt{[CLS]} token, which is trained to aggregate global image representation.
However, a growing number of modern vision encoders, such as SigLIP~\cite{Zhai:SigLIP}, have omitted this special token.
This design choice poses a challenge for methods like ours that leverage it.
To ensure the generalizability of FiCoCo-V, we explore the substitutability of the \texttt{[CLS]} token grounded in the mechanics of the self-attention mechanism.

\para{Theoretical justification: Why the mean of Key vectors?}
To find the most suitable proxy for a global context vector, we revisit the fundamental roles of the Query (Q), Key (K), and Value (V) vectors within the self-attention mechanism~\cite{Vaswani:Transformer}.

\begin{itemize}
\item \textbf{Query (Q)}: A token's Query vector acts as a ``probe'' or ``request'', signifying the information it seeks from other tokens in the sequence. The average of all Query vectors would thus represent the average information need of the image patches, not their collective content.
\item \textbf{Key (K)}: A token's Key vector serves as its ``identifier'' or ``advertisement''. It is the representation that the token ``broadcasts'' to the sequence, against which other tokens' Queries are matched. Consequently, the average of all Key vectors logically synthesizes a representation of the average content or ``gist'' of the entire set of visual tokens. This is conceptually analogous to the function of a \texttt{[CLS]} token.
\item \textbf{Value (V)}: A token's Value vector contains the actual information or ``payload'' that is transmitted to other tokens once attention scores are computed. Averaging Value vectors would yield a representation of the average information payload, which may be less discriminative for identifying semantically distinct regions than the average Key.
\end{itemize}

Based on this analysis, we hypothesize that the mean of all patch tokens' Key vectors is the most theoretically sound and effective proxy for a global context vector in the absence of a \texttt{[CLS]} token. It directly captures the aggregated semantic identity of the visual patches.

\begin{table}[!t]
\centering
\setlength\tabcolsep{1pt} %
\scalebox{0.96}{ %
    \begin{tabular}{l ccccc c} 
    \textbf{Method} & \textbf{VQA\textsuperscript{T}} & \textbf{MMB} & \textbf{POPE} & \textbf{MM-Vet} & \textbf{Vizwiz} & \textbf{Avg (\%)} \\
    \hline
    \multicolumn{7}{c}{\footnotesize \textit{TFLOPs=8.5}} \\
    LLaVA-1.5 & 58.2 & 66.1 & 86.4 & 31.6 & 50.0 & 58.46 \\
    \hline
    \multicolumn{7}{c}{\footnotesize \textit{TFLOPs=1.5$_{(\downarrow82.4\%)}$}} \\
    $\mathbf{a}_i^{\texttt{CLS}}$ & 55.5 & 60.2 & 79.8 & 26.8 & 52.4 & 54.94 \\
    $\mathbf{a}_i^{\texttt{H}}$ & 54.2 & 59.6 & 81.4 & 25.9 & 49.8 & 54.18 \\
    $\mathbf{a}_i^{\texttt{Eq}}(Quary)$ & 52.0 & 57.8 & 79.6 & 25.1 & 49.9 & 52.89 \\
    $\mathbf{a}_i^{\texttt{Eq}}(Value)$ & 54.3 & 61.4 & 81.0 & 25.4 & 50.8 & 54.59 \\
    $\mathbf{a}_i^{\texttt{Eq}}(Key)$ & 54.8 & 60.3 & 81.4 & 26.5 & 50.9 & 54.78 \\
    \hline
    \end{tabular}%
}
\vspace{-2mm}
\caption{\textbf{Comparison results across different benchmarks.} The evaluation includes multiple datasets and varying FLOPs settings.}
\vspace{-2mm}
\label{tab:non-cls} 
\end{table}

\para{Empirical validation.}
We empirically validate this theoretical choice through both qualitative and quantitative analyses.
We utilize attention mechanism vectors—Query (Q), Key (K), and Value (V)—along with feature representations (Hidden States) as baselines to compute global mean vectors. Equivalent token is then generated based on these vectors and subsequently used for experimental evaluation. The qualitative experimental results are shown in the Figure~\ref{fig:ana_qkv}.

The findings indicate that as the layer increases, the saliency of instruction-related features becomes more pronounced. When using Q, V, and feature vectors as baselines, as the visual feature encoding is completed (\textit{i.e.}, at layer 24), although the features in answer-relevant regions become more prominent, answer-irrelevant regions still exhibit a certain degree of saliency. This suggests that the redundant tokens selected based on these methods are more likely to overlap with the answer-relevant regions, potentially affecting the final information selection process.

In contrast, when using the K vector as the baseline, the mean token exhibits distinct characteristics. Although its saliency in answer-relevant regions is less pronounced compared to the Q, K, and feature vector baselines, the scores of answer-irrelevant regions are better suppressed. This implies that the influence of answer-irrelevant regions on relevant regions is reduced, allowing for more effective filtering of redundant tokens. As a result, this setting proves to be more efficient in preserving information critical to the final task.
Therefore, we compute the mean of the keys across the attention head dimension to mitigate local attention biases. 

Second, the quantitative results in Table~\ref{tab:non-cls} further corroborate our choice. 
We evaluate different proxies under an extreme compression setting (82.4\% TFLOPs reduction).
The results indicate that replacing the $\mathbf{a}_i^{\texttt{CLS}}$ with the $\mathbf{a}_i^{\texttt{Eq}}$ leads to only a 0.16\% decrease in average accuracy, demonstrating the effectiveness of the selected alternative token. Moreover, compared to directly obtaining an equivalent token using the mean of feature vectors ($\mathbf{a}_i^{\texttt{H}}$), the equivalent token computed based on keys achieves a 0.60\% improvement in average accuracy. Additionally, in comparison with equivalent tokens derived from Q and V, the key-based equivalent token improves average accuracy by 1.89\% and 0.19\%, respectively. These quantitative experimental results suggest that the equivalent token computed using key vectors can more comprehensively capture the 
global semantic information, thereby contributing to better performance.

\para{Implementation of the Equivalent Token.}
The specific implementation process of equivalent tokens is as follows.%
First, to mitigate bias from any single attention head, we compute the mean of the Key States across all $H$ attention heads:
\begin{equation}
    M = \frac{1}{H} \sum_{h=1}^{H} K_h
\end{equation}
where \( K_h \in \mathbb{R}^{B \times T \times D} \) represents the key States of the \( h \)-th attention head, \( H \) is the number of attention heads, and \( M \in \mathbb{R}^{B \times T \times D} \) represents the mean key.
To compute the patch tokens, we extract all tokens except the first one (\texttt{[CLS]} token):
\begin{equation}
    P = \{ M_i \}_{i=2}^{T} = \left[ M_2, M_3, \dots, M_T \right]
\end{equation}
where \( P \in \mathbb{R}^{B \times (T-1) \times D} \) contains the patch tokens, \( M_i \) represents the \( i \)-th token in \( M \), \( T-1 \) represents the number of patch tokens after removing the \texttt{[CLS]} token. Then the mean patch token is computed as:
\begin{equation}
    \mu = \frac{1}{T-1} \sum_{i=2}^{T} M_i
\end{equation}
where \( \mu \in \mathbb{R}^{B \times 1 \times D} \) represents the average patch token.

The cosine similarity between the mean patch token \( \mu \) and each patch token \( P_i \) is given by:
\begin{equation}
    \text{cos\_sim}(\mu, P_i) = \frac{\mu \cdot P_i}{\|\mu\|_2 \|P_i\|_2}
\end{equation}
where  \( \|\mu\|_2 \) is the L2 norm of the mean patch token, \( \|P_i\|_2 \) is the L2 norm of the \( i \)-th patch token. 

The final computed is
\begin{equation}
    \mathbf{a}_i^{\texttt{Eq}} = - \text{cos\_sim}(\mu, P_i)
\end{equation}

which results in a tensor of shape \( \mathbb{R}^{B \times (T-1)} \), representing the negative cosine similarity between the mean patch token and each individual patch token. The core reason for taking the negative of $\text{cos\_sim}(\mu, P_i)$ is that we aim to emphasize \textbf{the difference from the global mean vector} rather than its similarity. When $\text{cos\_sim}(\mu, P_i)$ is negative, it indicates that $\mu$ and $P_i$ are in opposite directions, suggesting that the token possesses a high degree of independence and contains crucial information. Conversely, when $\text{cos\_sim}(\mu, P_i)$ is positive, it implies that the token's features closely resemble the global mean, making it more likely to be redundant. By taking the negative, we prioritize preserving tokens with higher information density while suppressing the influence of redundant tokens, leading to a more precise selection of relevant information. Finally, by replacing $\mathbf{a}_i^{\texttt{CLS}}$ in Eq.~\ref{eq3} with $\mathbf{a}_i^{\texttt{Eq}}$, the \textit{FiCoCo-V} can be used in a version that does not require the \texttt{[CLS]} token.

\begin{figure}[!t]
  \centering
   \includegraphics[width=\linewidth]{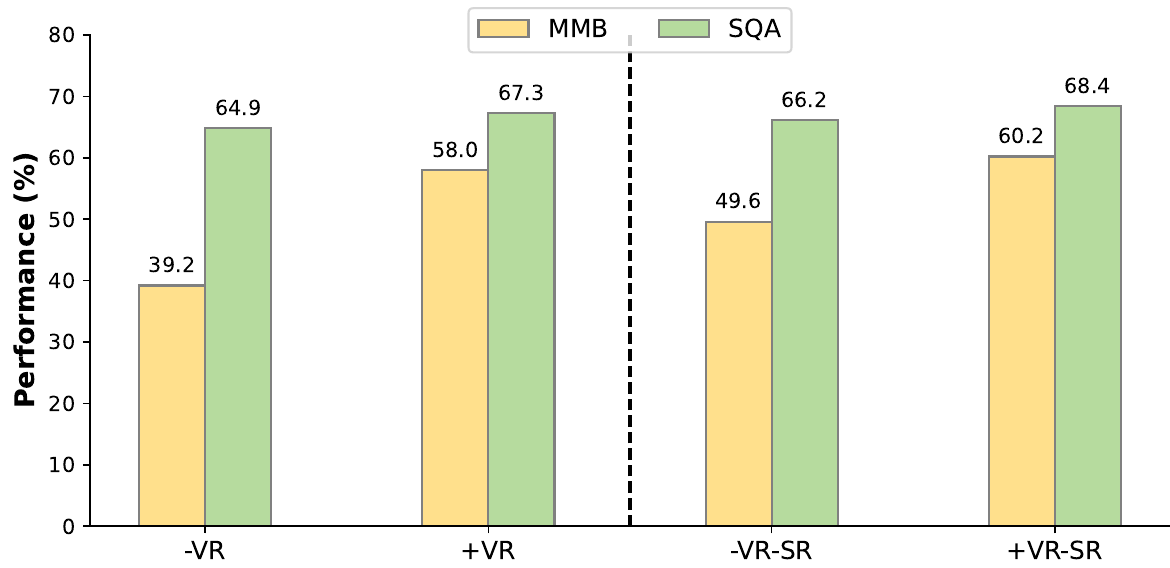}
   \caption{
   \textbf{Comparative experiment on the sign of vision-aware redundancy.} VR denotes vision-aware redundancy, while SR denotes semantic-aware redundancy. Our proposed formulation (+VR) shows superior performance.}  %
   
   \label{fig:VR-sign}
   \vspace{-4mm}
\end{figure}

\begin{figure}[!t]
  \centering
   \includegraphics[width=\linewidth]{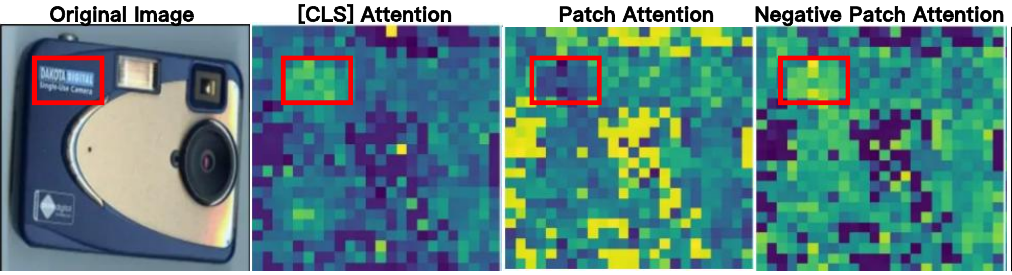}
   \caption{\textbf{Visualization of attention scores.} Brighter regions indicate higher attention weights, and red boxes highlight regions relevant to the answer. The ``Negative Patch Attention'' map (our redundancy metric) better aligns with the semantically focused ``\texttt{[CLS]} Attention'' map.}  %
   
   \label{fig:att_vis}
   \vspace{-4mm}
\end{figure}

\subsection{Analysis of the Sign of Vision-aware Redundancy}  \label{sec:sign-vision-aware-redundancy}

When filtering out the discarded tokens, we propose a core hypothesis grounded in an information-theoretic perspective. Our hypothesis is that the total attention a visual token receives from other visual tokens (\textit{i.e.}, its ``attention-in'' score) serves as a proxy for its predictability. In information theory, a signal that is highly predictable by its context has low information entropy and thus contains less ``new'' or unique information. Such a token is, by definition, redundant.

However, this perspective stands in stark contrast to prior works like FastV~\cite{Chen:FastV}, which interpret attention scores as a direct measure of importance. We argue that such approaches may conflate a token's importance with its predictability. A token might receive high attention because its content can be easily reconstructed from its neighbors (making it redundant), not necessarily because it is uniquely vital.
Here, we provide empirical evidence to validate this hypothesis.
First, we conduct a direct comparative experiment. We define our standard vision-aware redundancy term as \textbf{+VR} (as in Eq. 1, where it contributes positively to the total redundancy score). We then test an alternative formulation, \textbf{-VR}, where we flip the sign, aligning it with the ``attention as importance'' paradigm. This term is combined with our standard semantic-aware redundancy term (\textbf{-TR}). As shown in \Cref{fig:VR-sign}, the \textbf{+VR} configuration, which treats high inter-patch attention as redundancy, consistently and significantly outperforms the \textbf{-VR} configuration across multiple benchmarks. This result provides strong quantitative support for our hypothesis, demonstrating that penalizing predictable tokens is more effective for performance preservation than preserving them.

Second, we offer qualitative insight through visualization in \Cref{fig:att_vis}.
We compare the attention map of the \texttt{[CLS]} token (representing global semantic importance) with the averaged attention map of all patch tokens (representing inter-patch dependency, i.e., our vision-aware redundancy). The \texttt{[CLS]} token clearly focuses on the semantically critical regions relevant to the answer (the text on the bottle). In contrast, the patch-to-patch attention is more diffuse and highlights regions of textural similarity and local context. When we treat this patch attention as redundancy (visualized as ``Negative Patch Attention'', where brighter means less redundant), the resulting saliency map aligns much better with the \texttt{[CLS]} attention map. This visualization corroborates our information-theoretic view: tokens that are less predictable by their peers (darker in the ``Patch Attention'' map) are the ones that carry the most unique, globally relevant information.

Together, these quantitative and qualitative results provide strong validation for our core hypothesis, establishing a more principled foundation for identifying and filtering redundant visual tokens.

\begin{figure}[!t]
  \centering
   \includegraphics[width=\linewidth]{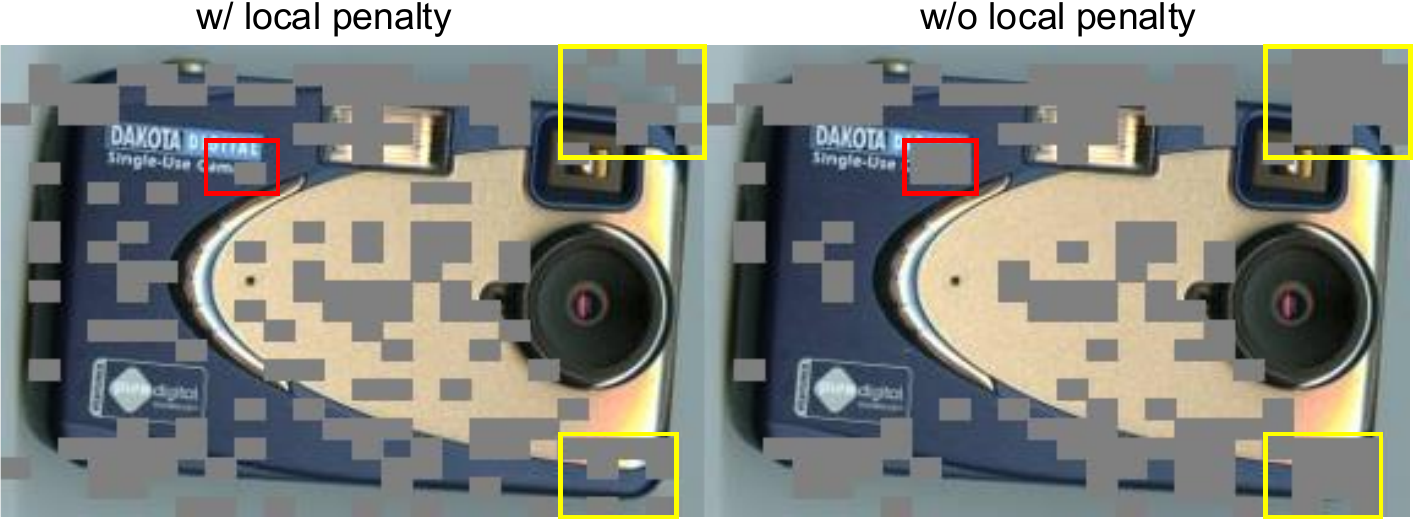}
   \caption{\textbf{Visualization of local penalty.} Yellow boxes denote regions irrelevant to the answer, whereas red boxes indicate answer-relevant areas.}  %
   \label{fig:local_penalty}
   \vspace{-2mm}
\end{figure}

\subsection{Analysis of Local Penalty Strategy}
We conduct qualitative analyses to validate the effectiveness of the proposed local penalty strategy. In the qualitative study, we retain 200 visual tokens and evaluate the spatial uniformity of the discarded tokens. \Cref{fig:local_penalty} demonstrate that introducing the local penalty strategy significantly enhances the spatial uniformity of token distribution. Specifically, as highlighted in the red-boxed regions of the figure, without this strategy, token compression tends to concentrate spatially, leading to substantial information loss in critical regions and causing irreversible degradation in the final model output. In contrast, the local penalty strategy mitigates this global averaging tendency, promoting a more uniform spatial distribution of compressed tokens and better preservation of key information.

\begin{table}[!t]
  \centering
\tablestyle{5pt}{1.0}
\setlength\tabcolsep{1pt} %
\def\w{20pt} 
\scalebox{0.92}{
    \begin{tabular}{l|cccc}
    \textbf{Method} & \textbf{Quant} & \textbf{TFLOPs↓} & \textbf{Memory (GB)↓} & \textbf{KV-Cache (MB)↓} \\
    \shline
    LLaVA-1.5 & FP16  & 8.5   & 22.4  & 333 \\
    \textit{\textbf{FiCoCo-V}} & FP16  & 1.5 \mytiny{(↓82\%)}  & 14.4 \mytiny{(↓36\%)} & 65.0 \mytiny{(↓80\%)} \\
    \textit{\textbf{FiCoCo-L}} & FP16  & 1.5 \mytiny{(↓82\%)}  & 14.3 \mytiny{(↓36\%)} & 64.2 \mytiny{(↓81\%)} \\
    \hline
    LLaVA-1.5 & INT8  & 4.3   & 11.2  & 167 \\
    \textit{\textbf{FiCoCo-V}} & INT8  & 0.8 \mytiny{(↓81\%)}  & 7.8 \mytiny{(↓30\%)}   & 32.5 \mytiny{(↓81\%)} \\
    \textit{\textbf{FiCoCo-L}} & INT8  & 0.8 \mytiny{(↓81\%)}  & 7.2 \mytiny{(↓36\%)}  & 32.1 \mytiny{(↓81\%)} \\
    \hline
    LLaVA-1.5 & INT4  & 2.1   & 6.2   & 83.4 \\
    \textit{\textbf{FiCoCo-V}} & INT4  & 0.4 \mytiny{(↓81\%)}  & 4.4 \mytiny{(↓29\%)}  & 16.3 \mytiny{(↓81\%)} \\
    \textit{\textbf{FiCoCo-L}} & INT4  & 0.4 \mytiny{(↓81\%)}  & 3.3 \mytiny{(↓47\%)}  & 16.1 \mytiny{(↓81\%)} \\
    \end{tabular}%
}
    \caption{Efficiency analysis of methods based on LLaVA-1.5-7B.}
  \label{tab:efficieny-7B}%
  \vspace{-2mm}
\end{table}%

\begin{table}[h]
  \centering
\tablestyle{5pt}{1.0}
\setlength\tabcolsep{2.5pt} %
\def\w{20pt} 
\scalebox{0.9}{
    \begin{tabular}{l|cccc}
    \textbf{Method} & \textbf{Quant} & \textbf{TFLOPs↓} & \textbf{Memory (GB)↓} & \textbf{KV-Cache (MB)↓} \\
    \shline
    LLaVA-1.5 & FP16  & 28.6  & 56.1  & 891 \\
    \textit{\textbf{FiCoCo-V}} & FP16  & 15.4 \mytiny{(↓46\%)} & 38.6 \mytiny{(↓31\%)} & 488 \mytiny{(↓43\%)} \\
    \textit{\textbf{FiCoCo-L}} & FP16  & 15.4 \mytiny{(↓46\%)} & 38.4 \mytiny{(↓32\%)} & 485 \mytiny{(↓46\%)} \\
    \hline
    LLaVA-1.5 & INT8  & 14.3  & 28   & 446 \\
    \textit{\textbf{FiCoCo-V}} & INT8  & 7.7 \mytiny{(↓46\%)}  & 19.3 \mytiny{(↓31\%)} & 244 \mytiny{(↓45\%)} \\
    \textit{\textbf{FiCoCo-L}} & INT8  & 7.7  \mytiny{(↓46\%)} & 19.2 \mytiny{(↓31\%)} & 242 \mytiny{(↓46\%)} \\
    \hline
    LLaVA-1.5 & INT4  & 7.6   & 14    & 223 \\
    \textit{\textbf{FiCoCo-V}} & INT4  & 3.9 \mytiny{(↓46\%)}  & 9.6 \mytiny{(↓32\%)}  & 122 \mytiny{(↓49\%)} \\
    \textit{\textbf{FiCoCo-L}} & INT4  & 3.9  \mytiny{(↓49\%)} & 9.5 \mytiny{(↓32\%)}  & 121 \mytiny{(↓46\%)} \\
    \end{tabular}%
}
    \caption{Efficiency analysis of methods based on LLaVA-1.5-13B.}
  \label{tab:efficieny-13B}%
  \vspace{-2mm}
\end{table}%

\begin{figure}[!t]
  \centering
   \includegraphics[width=\linewidth]{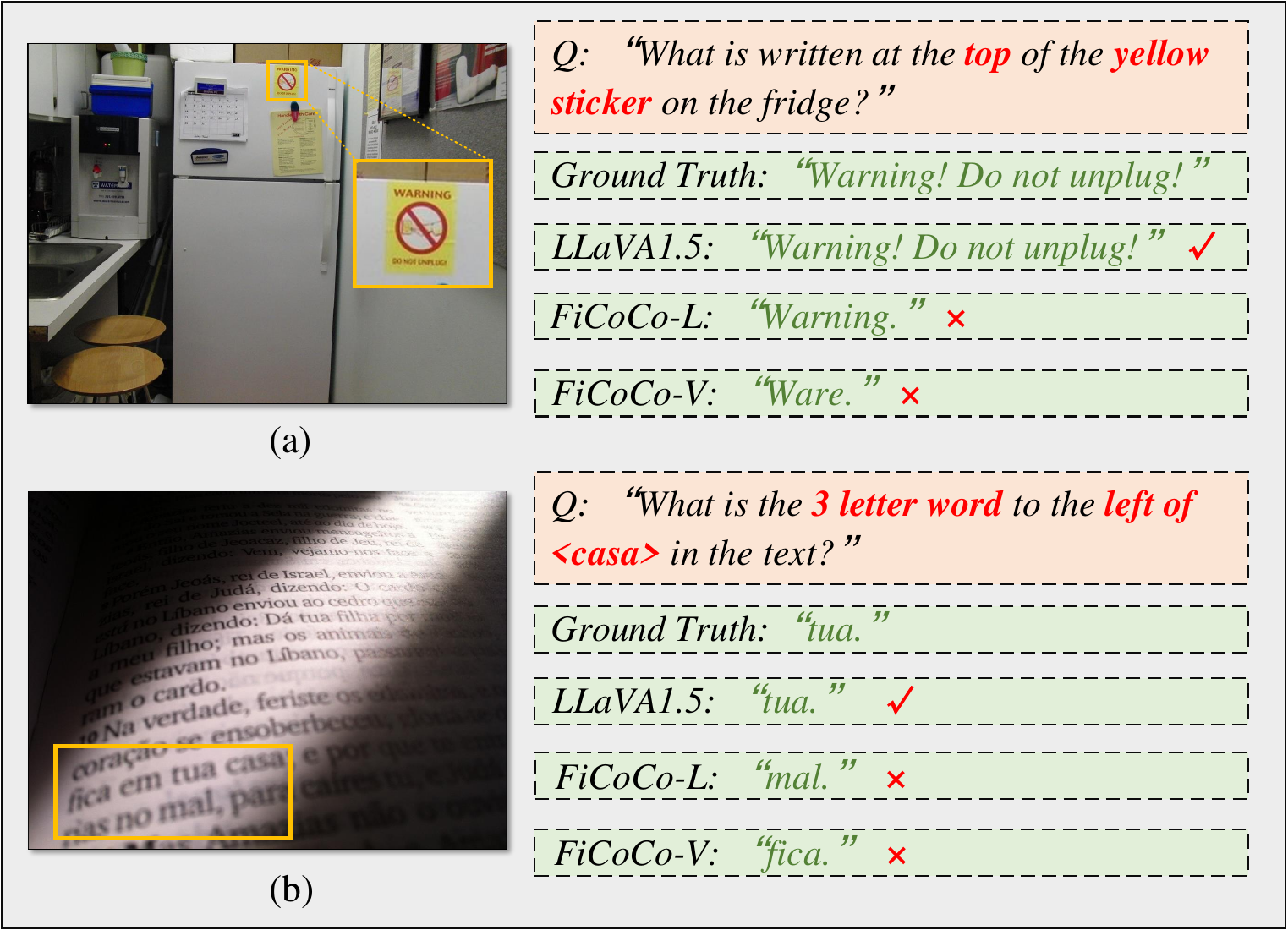}
   \caption{\textbf{Failure cases of \textit{FiCoCo}}. \textit{FiCoCo-L} produces answers more closely aligned with the questions.}  %
   
   \label{fig:Failure Case}
   \vspace{-4mm}
\end{figure}

\subsection{Detailed Efficiency Analysis}

Utilizing the tools provided by~\cite{Yuan:LLM-Viewer}, we conduct a detailed analysis of the theoretical efficiency of our \textit{FiCoCo}.
In \Cref{tab:efficieny-7B}, we assume the number of textual tokens is 60 for LLaVA-1.5-7B.
And in \Cref{tab:efficieny-13B}, we assume the number of textual tokens is 512 for LLaVA-1.5-13B.
The results demonstrate that, compared to the baseline models of LLaVA-1.5-7B/13B, our \textit{FiCoCo} series achieve significant improvements in both computational efficiency and GPU memory utilization. Specifically, our \textit{FiCoCo} series reduces computational overhead by nearly 80\%, GPU memory usage by approximately 40\%, and KV-Cache storage by around 80\%, all while achieving performance comparable to LLaVA-1.5-7B. Notably, this is accomplished without requiring any additional training, highlighting the efficiency and flexibility of our \textit{FiCoCo} series.

\begin{figure*}[!t]
  \centering
   \includegraphics[width=\linewidth]{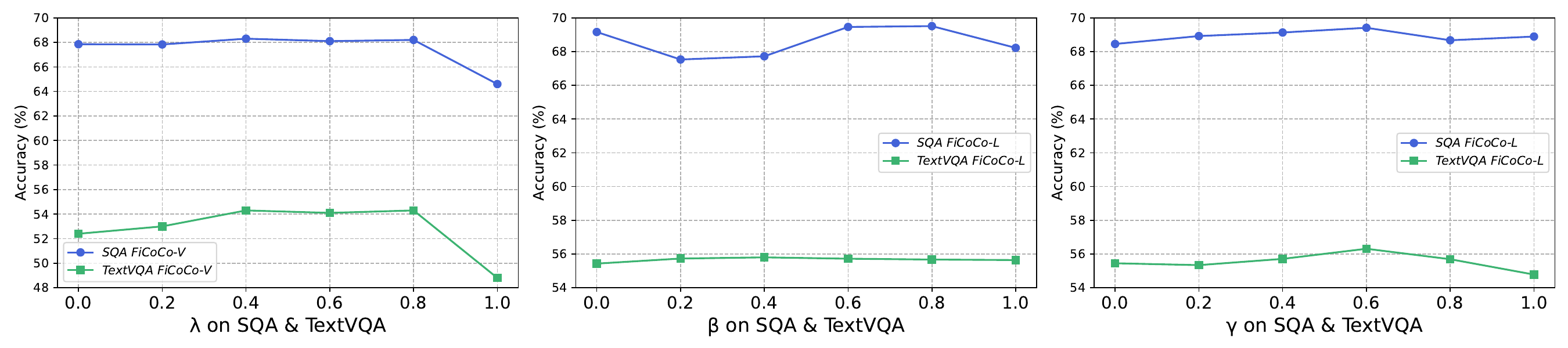}
   \caption{Hyperparameter sensitivity analysis of $\lambda$, $\beta$ and $\gamma$ on TextVQA and SQA benchmarks.}  %
   
   \label{fig:hyper_para}
   \vspace{-4mm}
\end{figure*}

\subsection{Analysis of Failure Cases}

\textit{FiCoCo} maintains substantial performance even when compressing a significant number of visual tokens.
However, the inevitable loss of visual information during the token reduction still causes failure cases.
We show two cases in \Cref{fig:Failure Case} where the answers generated by LLaVA-1.5 are consistent with the ground truth, while \textit{FiCoCo-L} and \textit{FiCoCo-V} 
fail to answer correctly.
By analyzing the erroneous responses generated by \textit{FiCoCo-L} and \textit{FiCoCo-V}, it can be observed that \textit{FiCoCo-L} produces answers more closely aligned with the questions, guided by the token selection process involving textual information.
For instance, in \Cref{fig:Failure Case}(a), the prompts `\textit{top}' and `\textit{yellow sticker}' jointly indicate the yellow region at the top of the refrigerator, leading \textit{FiCoCo-L} to search for the answer in this specific region. However, \textit{FiCoCo-V} fails to attend to the crucial information regarding `\textit{top}'.
Moreover, in \Cref{fig:Failure Case}(b), the cues `\textit{3 letter word}' and `\textit{left of casa}' jointly guide the answer towards `\textit{tua}.' Although the generated answer of \textit{FiCoCo-L} is `\textit{mal}', it more effectively considers these two cues. In contrast, \textit{FiCoCo-V} fails to adequately track the critical information pertaining to `\textit{3 letter word}.'

\begin{table}[!t]
  \centering
\tablestyle{5pt}{1.0}
\setlength\tabcolsep{3.5pt}
\def\w{20pt} 
\scalebox{1}{
    \begin{tabular}{c|cccc}
          & \multicolumn{2}{c}{\textit{FiCoCo-V}} & \multicolumn{2}{c}{\textit{FiCoCo-L}} \\
    \textbf{$\varepsilon$} & \textbf{SQA} & \textbf{TextVQA} & \textbf{SQA} & \textbf{TextVQA} \\
    \shline
     0.998 & 68.37 & \textbf{55.46} & 69.46 & \textbf{55.72} \\
    0.996 & 68.33 & 53.15 & \textbf{69.51} & 55.62 \\
    0.994 & 68.21 & 52.05 & 69.32 & 55.42 \\
    0.992 & \textbf{68.47} & 52.29 & 69.36 & 55.14 \\
    \end{tabular}%
    }
    \caption{
Hyperparameter sensitivity analysis of $\varepsilon$ on TextVQA and SQA benchmarks.}
  \label{tab:varepsilon}%
\end{table}%

\begin{table}[!t]
  \centering
\tablestyle{5pt}{1.0}
\setlength\tabcolsep{3.5pt}
\def\w{20pt} 
\scalebox{1}{
    \begin{tabular}{c|cc}
    \textbf{scaling coefficient} & \multicolumn{2}{c}{\textit{FiCoCo-V}} \\
    \textbf{in local penalty strategy} & \textbf{SQA} & \textbf{TextVQA} \\
    \shline
    1     & 68.12 & 53.24 \\
    2     & \textbf{68.37} & 55.46 \\
    3     & 68.21 & 55.04 \\
    4     & 68.11 & \textbf{55.49} \\
    \end{tabular}%
    }
    \caption{
Hyperparameter sensitivity analysis of scaling coefficient in local penalty strategy on TextVQA and SQA benchmarks.}
  \label{tab:scaling-coefficient}%
  \vspace{-4mm}
\end{table}%

\subsection{Sensitivity Analysis of Hyperparameters}\label{suppl:analysis}

We explore the hyperparameter configurations of \textit{FiCoCo}, performing sensitivity analysis on individual parameters to assess their impact. The experiments are conducted on both TextVQA and SQA benchmarks,
with TFLOPs at 1.5.

\para{Trade-off hyperparameters.}
It is observed that: 
(1) The hyperparameter $\lambda = 0.35$ is the optimal setting. Under this configuration, both \textit{FiCoCo-V} and \textit{FiCoCo-L} variants achieve relatively optimal accuracy. This indicates that when $\lambda = 0.35$, \textit{FiCoCo} effectively balances the local information conveyed by patch tokens with the global information carried by the \texttt{[CLS]} token, thereby enhancing the integration of visual features and the completeness of information.
(2) The hyperparameter $\beta = 0.6$ is the optimal setting. For the SQA dataset, \textit{FiCoCo-L} demonstrates a clear upward trend between $\beta = 0.4$ and $\beta = 0.6$, with a similar trend observed on the TextVQA dataset. This finding suggests that, under this parameter setting, an effective balance is achieved between textual information and the information conveyed by patch tokens.
(3) The hyperparameter $\gamma = 0.6$ is the optimal setting. \Cref{fig:hyper_para} clearly shows that \textit{FiCoCo-V} and \textit{FiCoCo-L} both reach their performance peaks at $\gamma = 0.6$ across the two benchmarks. This result suggests that incorporating semantic similarity more effectively guides the selection of the target set during the compress stage, thereby optimizing overall performance.

\para{$\varepsilon$ hyperparameter.}
\Cref{tab:varepsilon} compares the impact of different quantile thresholds $\varepsilon$-th.
Experimental results demonstrate that setting $\varepsilon$ to 0.998 yields optimal performance on both the TextVQA and SQA benchmarks.
However, as $\varepsilon$-th decreases, the information of a single token gets distributed across more tokens,
which leads to a noticeable performance drop in both benchmarks due to the excessive information fusion.

\para{Scaling coefficient hyperparameter in local penalty strategy.}
\Cref{tab:scaling-coefficient} shows that when the scaling coefficient exceeds 2,
the performance stably closes to optimal.
Therefore, to balance design simplicity and performance stability, we opt to fix the punishment coefficient at 2.

\section{Algorithm Illustration}\label{sec:algorithm}

We provide a detailed explanation of our \textit{FiCoCo-V} and \textit{FiCoCo-L} processes in Algorithm \ref{alg:FiCoCo-V} and Algorithm \ref{alg:FiCoCo-L}, respectively, to facilitate a clearer understanding of the methods we propose.

\begin{algorithm}
\caption{\textit{FiCoCo-V}}
\label{alg:FiCoCo-V}
\begin{algorithmic}[1]
\Require Input tokens $\mathbf{X} \in \mathbb{R}^{N \times D}$, attention score tensor $\mathbf{A}^v \in \mathbb{R}^{N \times N}$, \texttt{[CLS]} attention score vector $\mathbf{a}^{\texttt{CLS}} \in \mathbb{R}^{N}$, reduction factor $N^{\mathbb{S}} \in \mathbb{R}$, number of visual tokens $N \in \mathbb{R}$, hyperparameters $\lambda$, $\varepsilon \in [0,1]$ 
\Ensure Output tokens $\mathbf{X} \in \mathbb{R}^{(N-N^{\mathbb{S}}) \times D}$

\noindent \State \textbf{Stage 1: Filter}

\noindent \State Compute redundancy scores for all visual tokens:
{\setlength\abovedisplayskip{1mm}
\setlength\belowdisplayskip{1mm}
\[
\mathbf{s}^v_i = \lambda \frac{1}{N} \sum_{j=1}^N \mathbf{A}^v_{i,j} - (1 - \lambda) \mathbf{a}_i^{\texttt{CLS}}
\]
}

\noindent \State Apply \textit{local penalty} to $\mathbf{s}^v$

\noindent \State Identify source set $\mathbb{S} = \text{topK}(\mathbf{s}^v, N^{\mathbb{S}})$ that contains the indices of $N^{\mathbb{S}}$ discarded visual tokens

\noindent \State Identify target set $\mathbb{T}$ that contains the indices of $(N-N^{\mathbb{S}})$ preserved visual tokens

\noindent \State \textbf{Stage 2: Correlate}

\noindent \State Construct correlation matrix: 
{\setlength\abovedisplayskip{1mm}
\setlength\belowdisplayskip{1mm}
\[
\mathbf{C}^v_{i,j} = \mathbf{A}^v_{i,j}, \quad i \in \mathbb{S}, \ j \in \mathbb{T}
\]
}

\noindent \State \textbf{Stage 3: Compress}

\noindent \State Apply token-wise quantile-based thresholding: 
{\setlength\abovedisplayskip{1mm}
\setlength\belowdisplayskip{1mm}
\[
\tau_i = \text{quantile}(\mathbf{C}^v_{i, :}, \varepsilon)
\]
}

\noindent \State Compute token-adaptive topK correlations:
{\setlength\abovedisplayskip{1mm}
\setlength\belowdisplayskip{1mm}
\[
\mathbb{I}_j = \{i \in \mathbb{S} \text{ and } \mathbf{C}^v_{i,j} \geq \tau_i\}, \quad
\mathbb{J}_i = \{j \in \mathbb{T} \text{ and } \mathbf{C}^v_{i,j} \geq \tau_i\}
\]
}

\noindent \State Compute compression weights:
{\setlength\abovedisplayskip{1mm}
\setlength\belowdisplayskip{1mm}
\[
\alpha_{ij} = \frac{\mathbf{C}^v_{i,j}}{\sum_{j \in \mathbb{J}_i} \mathbf{C}^v_{i,j}}
\]
}

\noindent \State Update preserved tokens with self-preserving compression:
{\setlength\abovedisplayskip{1mm}
\setlength\belowdisplayskip{1mm}
\[
\mathbf{X}^{\mathbb{T}}_j \gets \frac{\mathbf{X}^{\mathbb{T}}_j + \sum_{i \in \mathbb{I}_j} \alpha_{ij} \mathbf{X}^{\mathbb{S}}_i}{1 + \sum_{i \in \mathbb{I}_j} \alpha_{ij}}
\]
}

\noindent \State Output tokens:
\[
\mathbf{X} \gets \mathbf{X} \setminus \mathbf{X}^\mathbb{S}
\]

\noindent \State \textbf{Return} $\mathbf{X}$

\end{algorithmic}
\end{algorithm}

\begin{algorithm}[h]
\caption{\textit{FiCoCo-L}}
\label{alg:FiCoCo-L}
\begin{algorithmic}[1]
\Require  Input tokens $\mathbf{X} \in \mathbb{R}^{(N+M) \times D}$, attention score tensor $\mathbf{A}^l \in \mathbb{R}^{(N+M) \times (N+M)}$, reduction factor $N^{\mathbb{S}} \in \mathbb{R}$, number of visual tokens $N \in \mathbb{R}$, number of textual tokens $M \in \mathbb{R}$, hyperparameters $\beta$, $\gamma$, $\varepsilon \in [0,1]$
\Ensure Output tokens $\mathbf{X} \in \mathbb{R}^{(N+M-N^{\mathbb{S}}) \times D}$

\noindent \State \textbf{Stage 1: Filter}

\noindent \State Compute redundancy scores for all visual tokens:
{\setlength\abovedisplayskip{1mm}
\setlength\belowdisplayskip{1mm}
\[
\mathbf{s}^l_i = \beta \frac{1}{N} \sum_{j=1}^N \mathbf{A}^l_{i,j} - (1 - \beta) \sum_{k=N+1}^{N+M} \mathbf{A}^l_{i,k}
\]
}

\noindent \State Identify source set $\mathbb{S} = \text{topK}(\mathbf{s}^v, N^{\mathbb{S}})$ that contains the indices of $N^{\mathbb{S}}$ discarded visual tokens

\noindent \State Identify target set $\mathbb{T}$ that contains the indices of $(N-N^{\mathbb{S}})$ preserved visual tokens

\noindent \State \textbf{Stage 2: Correlate}

\noindent \State Compute direct and indirect correlations:
{\setlength\abovedisplayskip{1mm}
\setlength\belowdisplayskip{1mm}
\[
\mathbf{C}^l_{i,j} = \gamma \mathbf{A}^l_{i,j} + (1 - \gamma) \sum_{k=N+1}^{N+M} \mathbf{A}^l_{i,k} \cdot \mathbf{A}^l_{k,j}
\]
}

\noindent \State \textbf{Stage 3: Compress}

\noindent \State Apply token-wise quantile-based thresholding: 
{\setlength\abovedisplayskip{1mm}
\setlength\belowdisplayskip{1mm}
\[
\tau_i = \text{quantile}(\mathbf{C}^l_{i, :}, \varepsilon)
\]
}

\noindent \State Compute token-adaptive topK correlations:
{\setlength\abovedisplayskip{1mm}
\setlength\belowdisplayskip{1mm}
\[
\mathbb{I}_j = \{i \in \mathbb{S} \text{ and } \mathbf{C}^l_{i,j} \geq \tau_i\}, \quad \mathbb{J}_i = \{j \in \mathbb{T} \text{ and } \mathbf{C}^l_{i,j} \geq \tau_i\}
\]
}

\noindent \State Compute compression weights:
{\setlength\abovedisplayskip{1mm}
\setlength\belowdisplayskip{1mm}
\[
\alpha_{ij} = \frac{\mathbf{C}^l_{i,j}}{\sum_{j \in \mathbb{J}_i} \mathbf{C}^l_{i,j}}
\]
}

\noindent \State Update preserved tokens with self-preserving compression:
{\setlength\abovedisplayskip{1mm}
\setlength\belowdisplayskip{1mm}
\[
\mathbf{X}^{\mathbb{T}}_j \gets \frac{\mathbf{X}^{\mathbb{T}}_j + \sum_{i \in \mathbb{I}_j} \alpha_{ij} \mathbf{X}^{\mathbb{S}}_i}{1 + \sum_{i \in \mathbb{I}_j} \alpha_{ij}}
\]
}

\noindent \State Output tokens:
\[
\mathbf{X} \gets \mathbf{X} \setminus \mathbf{X}^\mathbb{S}
\]

\noindent \State \textbf{Return} $\mathbf{X}$

\end{algorithmic}
\end{algorithm}

\end{document}